\documentclass[sigconf]{acmart}
\AtBeginDocument{%
  \providecommand\BibTeX{{%
    \normalfont B\kern-0.5em{\scshape i\kern-0.25em b}\kern-0.8em\TeX}}}

\usepackage{booktabs}
\usepackage{float}
\usepackage{multirow}
\usepackage{textcomp}
\usepackage{subfigure} 
\usepackage{caption}
\usepackage{subcaption}
\usepackage{bm}
\usepackage{appendix}
\usepackage{algorithm}
\usepackage{algorithmic}
\usepackage{makecell}
\usepackage{url}
\usepackage[labelformat=simple]{subcaption}
\usepackage{CJKutf8}
\usepackage{latexsym}
\usepackage{amsfonts}
\usepackage{amsmath}
\usepackage{multicol}
\usepackage{multirow}
\usepackage{xspace}
\usepackage{booktabs}
\usepackage{bbding}
\usepackage{array}
\usepackage{threeparttable}
\usepackage{tcolorbox}
\usepackage{tabularx}
\usepackage{enumitem}
\usepackage{xcolor,colortbl}
\usepackage{color}
\usepackage{setspace}
\usepackage{makecell}
\usepackage{listings}
\usepackage{xltabular}
\usepackage{tcolorbox}
\tcbuselibrary{breakable}
\usepackage{xcolor}
\usepackage{arydshln}
\usepackage{csquotes}
\usepackage{hyperref}
\usepackage{url}
\usepackage{CJKutf8}
\usepackage{svg}
\usepackage{bm}
\usepackage{graphics}
\usepackage{graphicx}
\usepackage{array}
\usepackage[english]{babel}
\usepackage{textcomp}
\usepackage{latexsym}
\usepackage{siunitx}  % 用于对齐数字
\usepackage[normalem]{ulem}
\useunder{\uline}{\ul}{}
\usepackage{array}
\usepackage{titletoc}
\usepackage{tikz}
\usetikzlibrary{tikzmark}
\usepackage{pifont}

\makeatletter
\newcommand*\myfontsize{%
  \@setfontsize\myfontsize{7}{9}%
}
\makeatother
\newcommand{\mytextbox}[2]{\tikzmarknode[draw=#1,thick,inner sep=1pt]{test}{\myfontsize #2}}
\definecolor{myred}{rgb}{0.7, 0.3, 0.0}
\definecolor{myblue}{HTML}{0a41b8}
\definecolor{mygreen}{HTML}{056b34}
\definecolor{mypurple}{HTML}{5d1e8b}

\newcommand{\green}[1]{\mytextbox{mygreen}{\textbf{\textcolor{mygreen}{#1}}}}
\newcommand{\purple}[1]{\mytextbox{mypurple}{\textbf{\textcolor{mypurple}{#1}}}}

\newcommand{\mybluefont}[1]{{\color{myblue}#1}}

\copyrightyear{2026}
\acmYear{2026}
\setcopyright{cc}
\setcctype{by-nc-nd}
\acmConference[WWW '26] {Proceedings of the ACM Web Conference 2026}{April 13--17, 2026}{Dubai, United Arab Emirates.}
\acmBooktitle{Proceedings of the ACM Web Conference 2026 (WWW '26), April 13--17, 2026, Dubai, United Arab Emirates}
\acmISBN{979-8-4007-2307-0/2026/04}
\acmDOI{10.1145/3774904.3792460}

\author{Xiaoxi Li}
\authornote{Work done during internship at Xiaohongshu Inc.}
\affiliation{
  \institution{Renmin University of China}
  \city{Beijing}
  \country{China}
}
\email{xiaoxi_li@ruc.edu.cn}

\author{Wenxiang Jiao}
\author{Jiarui Jin}
\affiliation{
  \institution{Xiaohongshu Inc.}
  \city{Beijing}
  \country{China}
}

\author{Guanting Dong}
\author{Jiajie Jin}
\affiliation{
  \institution{Renmin University of China}
  \city{Beijing}
  \country{China}
}

\author{Yinuo Wang}
\affiliation{
  \institution{Tsinghua University}
  \city{Beijing}
  \country{China}
}
\author{Hao Wang}
\affiliation{
  \institution{Xiaohongshu Inc.}
  \city{Beijing}
  \country{China}
}

\author{Yutao Zhu}
\affiliation{
  \institution{Renmin University of China}
  \city{Beijing}
  \country{China}
}

\author{Ji-Rong Wen}
\affiliation{
  \institution{Renmin University of China}
  \city{Beijing}
  \country{China}
}
\email{jrwen@ruc.edu.cn}

\author{Yuan Lu}
\affiliation{
  \institution{Xiaohongshu Inc.}
  \city{Beijing}
  \country{China}
}
\email{luyuan3@xiaohongshu.com}

\author{Zhicheng Dou}
\authornote{Corresponding author.}
\affiliation{
  \institution{Renmin University of China}
  \city{Beijing}
  \country{China}
}
\email{dou@ruc.edu.cn}

\begin{document}
\begin{CJK}{UTF8}{gbsn}

\renewcommand{\shortauthors}{Xiaoxi Li et al.}

\title{DeepAgent: A General Reasoning Agent with Scalable Toolsets}

\begin{abstract}
Large reasoning models have demonstrated strong problem-solving abilities, yet real-world tasks often require external tools and long-horizon interactions. Existing agent frameworks typically follow predefined workflows, which limit autonomous and global task completion. In this paper, we introduce \textbf{DeepAgent}, an end-to-end deep reasoning agent that performs autonomous thinking, tool discovery, and action execution within a single, coherent reasoning process. To manage long-horizon interactions, we introduce an autonomous memory folding mechanism that compresses past interactions into structured episodic, working, and tool memories, reducing error accumulation while preserving critical information. To teach general-purpose tool use efficiently and stably, we develop an end-to-end reinforcement learning strategy, namely ToolPO, that leverages LLM-simulated APIs and applies tool-call advantage attribution to assign fine-grained credit to the tool invocation tokens. Extensive experiments on eight benchmarks, including general tool-use tasks (ToolBench, API-Bank, TMDB, Spotify, ToolHop) and downstream applications (ALFWorld, WebShop, GAIA, HLE), demonstrate that DeepAgent consistently outperforms baselines across both labeled-tool and open-set tool retrieval scenarios. The code and demo are available at \mybluefont{\textbf{\url{https://github.com/RUC-NLPIR/DeepAgent}}}.
\end{abstract}


% \ccsdesc[500]{Computing methodologies~Reinforcement learning}


\maketitle
% \title{DeepAgent: Towards End-to-end Agentic Reasoning with Scalable and Pluggable Toolsets}

\begin{figure}[!t]
\centering
\includegraphics[width=1\linewidth]{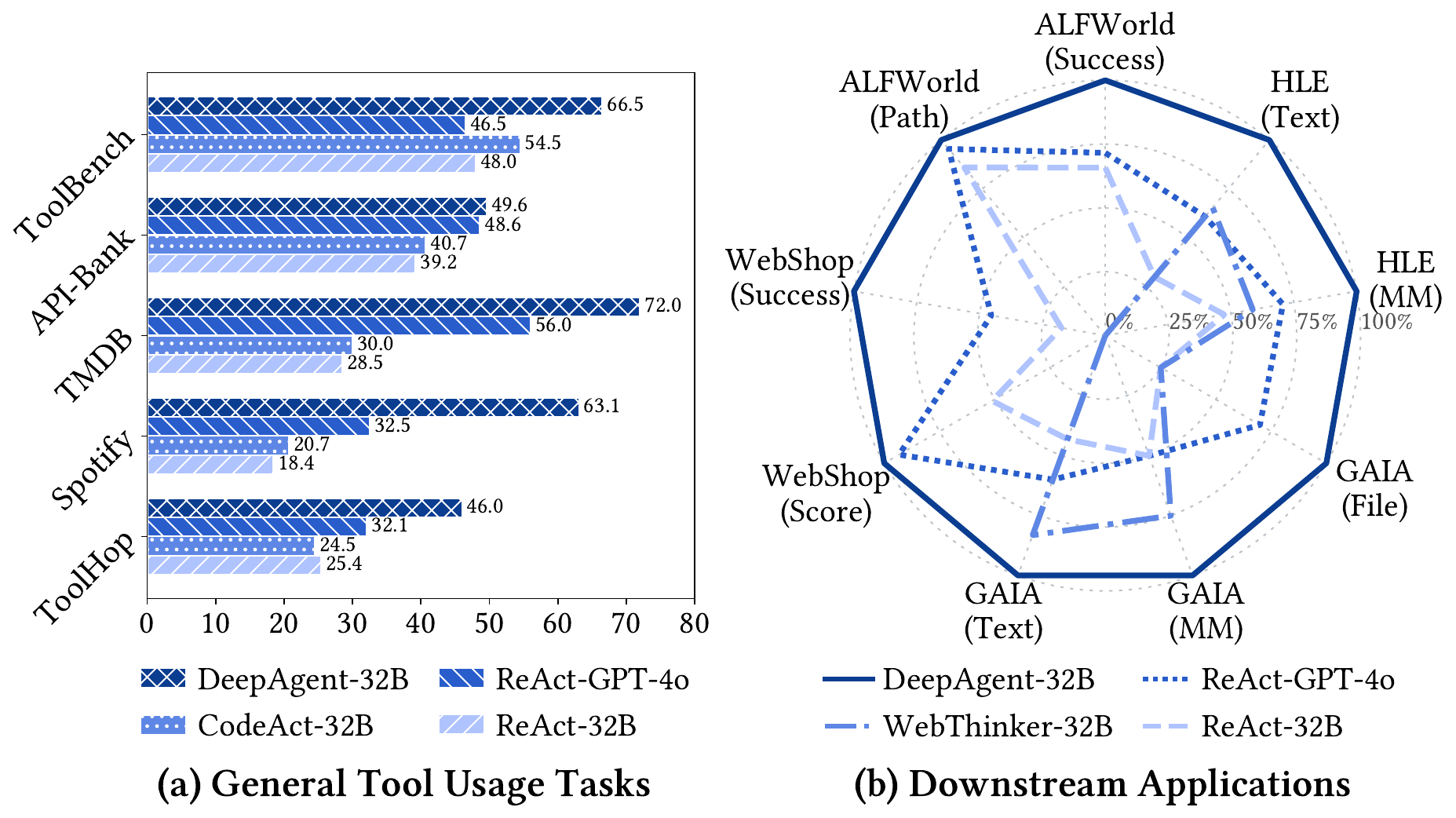}
\caption{Overall performance on (a) general tool usage tasks and (b) downstream applications (best score as 100\%).}
\label{fig:radar}
\end{figure}

\begin{figure*}[!t]
\centering
\includegraphics[width=0.925\linewidth]{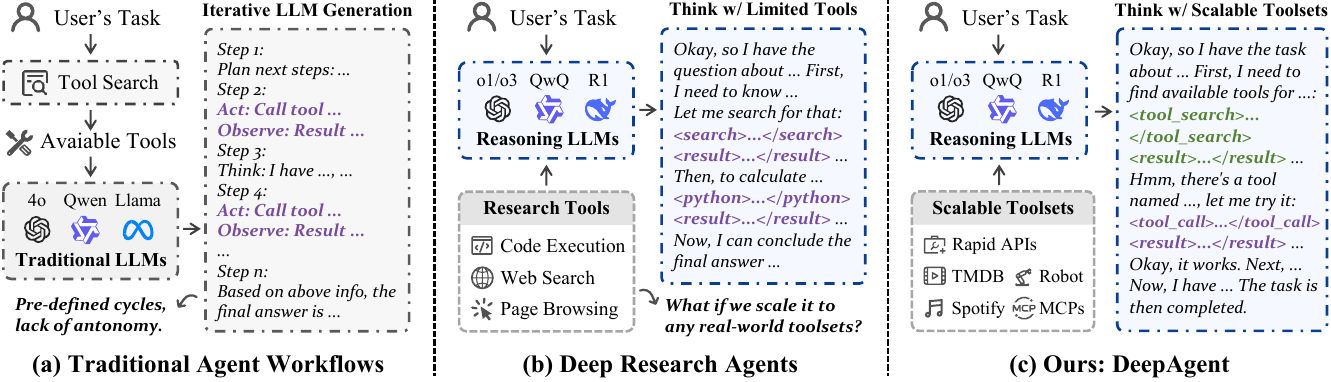}
\caption{
Comparison of agent paradigms: (a) Traditional agents with predefined workflows, (b) Deep Research agents that can autonomously call limited tools, and (c) Our DeepAgent, a fully autonomous reasoning agent that dynamically discovers and invokes helpful tools, all within a continuous agentic reasoning process. 
} 
\label{fig:compare}
\end{figure*}

\section{Introduction}

The rapid advancement of Large Language Models (LLMs) has inspired the development of LLM-powered agents, which have found broad applications in scenarios such as web information seeking, software engineering, and personal assistance~\cite{tool_learning_survey, llm_agent_se}. Most existing agents follow predefined workflows (e.g., ReAct~\cite{yao2022react} and Plan-and-Solve~\cite{Plan-and-Solve}) with iterative ``Reason-Act-Observe'' loops (Figure~\ref{fig:compare}(a)). Although effective in simpler tasks, these approaches suffer from several critical limitations: (1)~lack of autonomy in execution steps and overall procedure; (2)~inability to dynamically discover tools during task execution; (3)~deficiency in fully autonomous management of interactive memory; and (4)~insufficient depth and coherence in reasoning about the entire task. These limitations hinder agents from real-world problems, particularly for complex tasks that demand general and multiple tool-use.

Recently, the advent of Large Reasoning Models (LRMs) has demonstrated the capability to solve complex problems in domains like mathematics, programming, and scientific reasoning through a step-by-step ``slow thinking'' process~\cite{2503_Towards_Reasoning_Era, 2502_survey_from_system1, Video-Thinker}. However, many real-world tasks necessitate the use of external tools for their completion. Recent approaches integrate tool use into reasoning~\cite{2501_search_o1,2504_DeepResearcher,ToRL}, but typically rely on a small, fixed tool set such as search, browsing, and coding (Figure~\ref{fig:compare}(b)), limiting their generality.

To address these challenges, we introduce \textbf{DeepAgent}, an end-to-end deep reasoning agent that can complete an entire task by dynamically retrieving and calling tools within a single, coherent agentic reasoning process. As depicted in Figure~\ref{fig:compare}(c), DeepAgent operates by autonomously thinking, searching for tools, and executing actions. This paradigm shifts away from traditional, predefined workflows that rely on predefined tools, task planning, and iterative tool use. Instead, DeepAgent maintains a global perspective on the entire task, unconstrained by the need to deliberate on specific, isolated operations. Tools are not pre-retrieved in advance but are dynamically discovered on an as-needed basis, thereby fully unlocking the autonomous potential of the large reasoning model.

% To empower DeepAgent to thoroughly and robustly explore new tools and navigate complex environments during long-horizon interactions, we equip it with memory management capabilities. We introduce an \textbf{Autonomous Memory Folding} strategy that allows DeepAgent to consolidate its previous thoughts and interaction history into a \textit{structured memory schema} at any point during its thinking before resuming the agentic reasoning process. This mechanism not only saves tokens and enhances reasoning efficiency over extended interactions but also provides the agent an opportunity to ``take a breath'', preventing it from becoming trapped in wrong exploration paths and enabling it to reconsider its strategy. To mitigate information loss during folding, we introduce a \textit{brain-inspired memory architecture} comprising episodic memory, working memory, and tool memory, all structured with an agent-usable data schema to ensure the stability and utility of the folded memory.
To facilitate robust exploration in long-horizon environments, we equip DeepAgent with \textbf{Autonomous Memory Folding}. This strategy allows the agent to dynamically consolidate its reasoning process and interaction history into a \textit{structured memory schema}. Beyond reducing token overhead, this mechanism enables the agent to ``take a breath''—pausing to reconsider strategies and avoid erroneous paths. To minimize information loss during consolidation, we introduce a \textit{brain-inspired memory architecture} comprising episodic, working, and tool memory, all structured with an agent-usable data schema to ensure the stability and utility of the folded memory.

To enhance DeepAgent's proficiency in mastering these mechanisms, we propose \textbf{ToolPO}, an end-to-end reinforcement learning (RL) training method tailored for general tool use. 
Existing agentic RL training in general domains presents two significant challenges: (1)~The reliance on a multitude of real-world APIs during training can lead to instability, slow execution, and high costs. To prevent this, we leverage \textit{LLM-simulated APIs}, which enhance the stability and efficiency of the training process. (2)~A sparse reward based solely on the final outcome is often insufficient to guarantee the accuracy of intermediate tool calls. We address this by implementing \textit{tool-call advantage attribution}, which precisely assigns credit to the specific tokens responsible for correct tool invocations, thereby providing a more granular and effective learning signal.

We conduct extensive experiments on a wide range of benchmarks. For \textbf{(1)~General Tool-Use Tasks}, we evaluate DeepAgent on ToolBench, API-Bank, TMDB, Spotify, and ToolHop, which feature toolsets scaling from tens to over ten thousand distinct tools. For \textbf{(2)~Downstream Applications}, we test its performance on ALFWorld, WebShop, GAIA, and Humanity's Last Exam (HLE), which require the use of domain-specific toolsets. The overall results in Figure~\ref{fig:radar} show that DeepAgent achieves superior performance across all scenarios.

Our main contributions are summarized as follows:
\begin{enumerate}[leftmargin=1em]
\item We propose DeepAgent, the first agentic framework that enables reasoning models to autonomously think, discover tools, and execute actions within a unified reasoning process, empowering LRMs to harness toolsets of arbitrary scale and generalize to complex real-world tasks.
\item We introduce an autonomous memory folding mechanism, complemented by a brain-inspired memory design. This endows the agent with the ability to ``take a breath'' and reconsider its exploration strategies following unsuccessful attempts. %enhancing its robustness and adaptability.
\item We propose an end-to-end reinforcement learning training methodology for general-purpose tool use, ensuring stability and efficiency in large-scale tool execution during training, as well as accuracy in tool invocation during reasoning.
\item We conduct extensive experiments across eight benchmarks, demonstrating DeepAgent's superior tool-use capabilities and high adaptability to real-world tasks.
\end{enumerate}

\begin{figure*}[!t]
\centering
\setlength{\abovecaptionskip}{0.2cm}
\setlength{\belowcaptionskip}{0.2cm}
\includegraphics[width=0.93\linewidth]{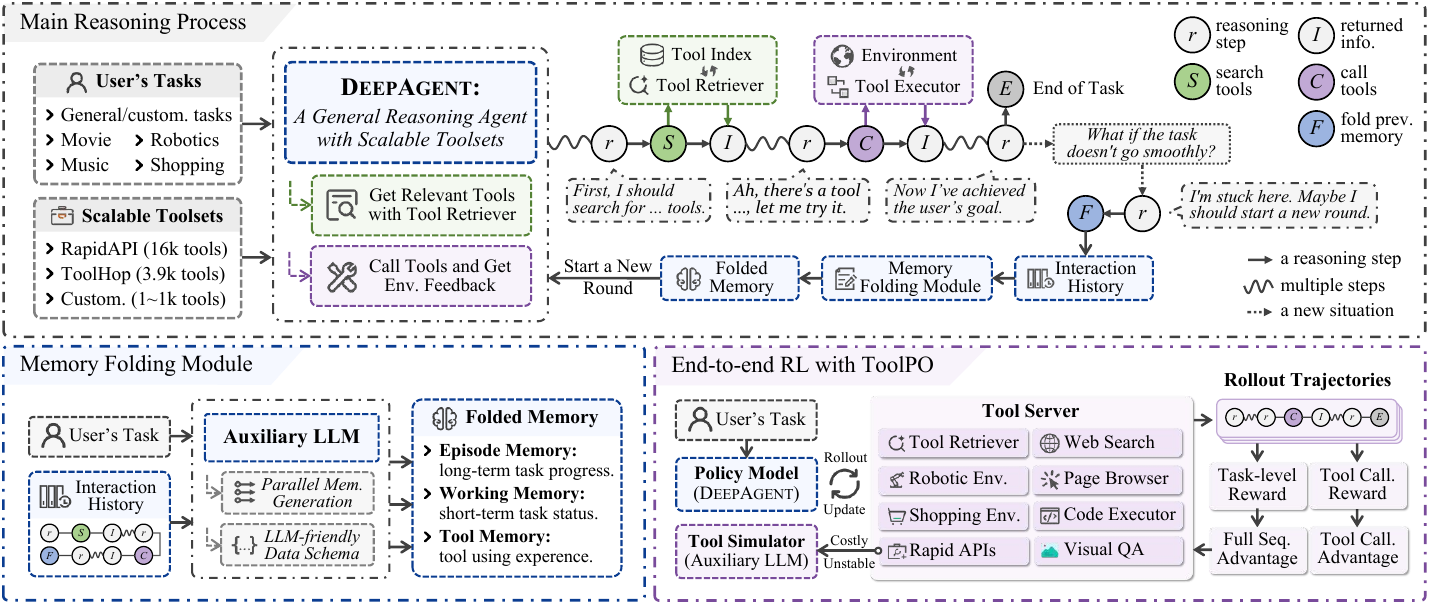}
\caption{
Overview of the DeepAgent framework. The main reasoning model autonomously discovers tools, executes actions, and folds previous memory to restart with structured memories, all within a unified thinking process. 
% When memory folding occurs, the system compresses interaction history into structured episodic, working, and tool memories. 
The DeepAgent is trained end-to-end with ToolPO, an RL method that uses a tool simulator to simulate large-scale real-world tool APIs, and rewards both final task success and correct intermediate tool calls through fine-grained advantage attribution.
}
\label{fig:architecture}
\end{figure*}

\section{Related Work}
\label{sec:related}

\subsection{Large Reasoning Models}
Large Reasoning Models (LRMs) \citep{deepseek-r1,openai2024openaio1card} have demonstrated significant performance improvements in mathematical, scientific, and coding tasks by employing step-by-step slow thinking processes before generating final responses. Existing research has explored various approaches to elicit extended Chain-of-Thought (CoT) reasoning~\cite{Wei_CoT} from models, including data synthesis for Supervised Fine-Tuning (SFT)~\cite{2410_o1_journey_part1, SyntheticCurriculum}, and end-to-end RL \citep{deepseek-r1}. 
Additionally, substantial work has investigated optimization strategies for reasoning models, such as advanced RL training algorithms~\cite{Diffusion-RL} and improving reasoning efficiency~\cite{Thinking-Optimal}. However, models relying solely on parametric knowledge face inherent limitations and cannot interact with the real world.
Recent studies have begun exploring tool-augmented reasoning approaches, including Search-o1 \citep{2501_search_o1}, Search-R1 \citep{2503_Search_R1}, ToRL \citep{ToRL}, DeepResearcher \citep{2504_DeepResearcher}, and SimpleTIR \citep{xue2025simpletir}. However, these methods typically support only a limited set of research-oriented tools, such as web search, page browsing, and code execution, which constrains their applicability to real-world scenarios that demand access to more diverse tools.

\subsection{Autonomous Agents}
LLM-powered autonomous agents accomplish real-world tasks by invoking external tools to interact with their environment~\cite{DRAFT, WebDancer, Tool-Star, WebExplorer, unigen, FinSight, RetroLLM, RAG-Critic, LongRefiner, HierSearch, TourPlanner, CompactClueSelection-RAG, EnvScaler, HiAgent, AgentGen, Agent2World}. 
Current agent methodologies, including ReAct~\cite{yao2022react}, Plan-and-Solve~\cite{Plan-and-Solve}, Reflextion~\cite{shinn2024reflexion}, and CodeAct~\cite{CodeAct}, predominantly follow predefined workflows with fixed execution patterns. This rigid structure limits their ability to fully leverage the autonomous decision-making and deep reasoning capabilities of advanced reasoning models.
Recent efforts have investigated training LLMs to autonomously invoke tools through data synthesis and SFT methods~\cite{AgentScaler, LIMI} and RL training frameworks~\cite{ReTool, RAGEN, VerlTool, GEM, ACON, FoldGRPO, ARPO, ToolLight, AEPO}. However, most existing methods rely on pre-selected, labeled tools, which limit their applicability to real-world scenarios. Real-world tasks are highly variable and require access to diverse toolsets that cannot be predetermined, aligning with the emerging Model Context Protocol (MCP)~\cite{MCP} paradigm. 
Although some prior work has explored tool retrieval mechanisms~\cite{ToolLLM, ToolGen, ToolRet}, most approaches conduct only a single upfront retrieval step and incorporate the retrieved tools, with limited exploration of dynamic tool discovery during task execution. Therefore, we aim to develop a deep reasoning agent capable of dynamically discovering and invoking helpful tools from scalable toolsets to address more generalized real-world tasks.

\section{Methodology}

% In this section, we first formulate the task of autonomous agentic reasoning. Then, we provide a detailed overview of the DeepAgent framework. Finally, we elaborate on the core components of DeepAgent, including the mechanism for autonomous tool use and memory folding, the brain-inspired memory schema, and our end-to-end reinforcement learning training method, ToolPO.

\subsection{Problem Formulation}

We frame the agent's task as a sequential decision-making process. The agent receives a user-provided question $Q$ and an instruction $I$, and interacts with an environment over a series of steps $t=1, \dots, T$ to accomplish the specified goal. The environment provides access to a collection of tools $\mathcal{T}$ at an arbitrary scale.

At each step $t$, the agent's state $s_t$ consists of the history of all previous actions and their resulting observations, i.e., $s_t = (a_1, o_1, \dots, a_{t-1}, o_{t-1})$. The agent, driven by a policy $\pi$ parameterized by $\theta$, selects an action $a_t$ based on the current state, the user question, and the instruction:
\begin{equation}
    a_t \sim \pi_\theta(\cdot | s_t, Q, I).
\end{equation}
An action $a_t$ can be one of four types:
\begin{itemize}[leftmargin=1em, nosep]
    \item \textbf{Internal Thought ($a_t^{\text{think}}$)}: A textual reasoning step generated by the LRM to analyze the problem or plan its next steps. The corresponding observation $o_t$ is typically empty.
    \item \textbf{Tool Search ($a_t^{\text{search}}$)}: A natural language query $q_s$ to find relevant tools from $\mathcal{T}$. The observation $o_t$ is a list of retrieved tools.
    \item \textbf{Tool Call ($a_t^{\text{call}}$)}: The invocation of a specific tool $\tau \in \mathcal{T}$ with a set of arguments. The observation $o_t$ is the execution result returned by the tool.
    \item \textbf{Memory Fold ($a_t^{\text{fold}}$)}: A special action to compress the interaction history $s_t$ into a structured memory summary. The subsequent state $s_{t+1}$ is then initialized with this compressed memory.
\end{itemize}

The sequence of states, actions, and observations forms a trajectory $\tau = (s_1, a_1, o_1, \dots, s_T, a_T, o_T)$. The process terminates when the agent completes the task or reaches a maximum step limit. Suppose $R(\tau)$ is a reward function that evaluates the overall success of the trajectory $\tau$, the objective is to learn an optimal policy $\pi_\theta^*$ that maximizes the expected cumulative reward for a given task:
\begin{equation}
    \pi_\theta^* = \arg\max_{\pi_\theta} \mathbb{E}_{\tau \sim \pi_\theta} [R(\tau)].
\end{equation}
%where $R(\tau)$ is a reward function that evaluates the overall success of the trajectory $\tau$.

\subsection{Overview of the DeepAgent Framework}

As illustrated in Figure~\ref{fig:architecture}, the DeepAgent framework is architected around a main reasoning process, which is supported by several auxiliary mechanisms to ensure robustness and efficiency.

\begin{itemize}[leftmargin=1em, nosep]
\item \textbf{Main Reasoning Process}: The core of DeepAgent is a powerful large reasoning model that drives the entire task-completion process. In a single stream of thought, the LRM autonomously reasons about the task, dynamically discovers necessary tools, executes actions, and manages its own memory. This unified approach departs from traditional, rigid agent workflows, allowing the LRM to maintain a global perspective on the task.

\item \textbf{Auxiliary Mechanisms}: DeepAgent employs an auxiliary LLM to handle complex interactions with large toolsets and manage long histories. This background model enhances system stability by: (1) filtering and summarizing retrieved tool documentation if it's too lengthy, (2) denoising and condensing verbose information returned from tool calls, and (3) compressing long interaction histories into a structured memory. This division of labor allows the main LRM to concentrate on high-level strategic reasoning.
\end{itemize}

\subsection{Autonomous Tool Search and Calling}
DeepAgent's main LRM performs all actions by generating specific textual prompts within its continuous reasoning process. These actions are then intercepted and executed by the system.

\paragraph{Tool Search}
When the agent determines it needs a tool, it generates a tool search query $q_s$ encapsulated within special tokens: \texttt{<tool\_search>} $q_s$ \texttt{</tool\_search>}. The system's tool retriever operates via dense retrieval. First, we build an index by pre-computing an embedding $E(d_i)$ for the documentation $d_i$ of each tool $\tau_i \in \mathcal{T}$ using an embedding model $E$. During inference, given the query $q_s$, the system retrieves the top-$k$ tools by ranking them based on the cosine similarity $\text{sim}(\cdot, \cdot)$:
\begin{equation}
\mathcal{T}_{\text{retrieved}} = \underset{\tau_i \in \mathcal{T}}{\text{top-k}} \left( \text{sim}\left(E\left(q_s\right), E\left(d_i\right)\right) \right).
\end{equation}
The retrieved tool documentation is then processed by the auxiliary LLM ---summarized if too lengthy, otherwise provided directly---and returned to the main LRM's context: \texttt{<tool\_search\_result>} relevant tools \texttt{</tool\_search\_result>}.

\paragraph{Tool Call}
To execute a tool, the agent generates a structured call including the tool's name and arguments: \texttt{<tool\_call>} \{"name": "tool\_name", "arguments": ...\} \texttt{</tool\_call>}. The framework parses this call, executes the tool, and captures the output. This output is, if necessary, summarized by the auxiliary LLM to ensure it is concise and helpful, before being fed back into the reasoning context: \texttt{<tool\_call\_result>} helpful information \texttt{</tool\_call\_result>}.

\subsection{Autonomous Memory Folding and Brain-Inspired Memory Schema}
The agent can trigger memory folding at any logical point in its reasoning process—such as after completing a sub-task or realizing an exploration path was incorrect—by generating a special token: \texttt{<fold\_thought>}. 
Upon detecting this token, the system initiates the memory folding process. The auxiliary LLM (parameterized by $\theta_{\text{aux}}$) processes the entire preceding interaction history $s_t$ and generates three structured memory components in parallel:
\begin{equation}
    (M_E, M_W, M_T) = f_{\text{compress}}(s_t; \theta_{\text{aux}}).
\end{equation}
These compressed episodic ($M_E$), working ($M_W$), and tool ($M_T$) memories then replace the raw interaction history, enabling the agent to proceed with a refreshed and condensed view of its progress while avoiding entrapment in incorrect exploration paths.

Inspired by human cognitive systems, the structured memory $M_t$ is composed of three distinct components that are generated in parallel: $M_t = (M_E, M_W, M_T)$, where $M_E, M_W, M_T$ denote episodic, working, and tool memories, respectively.

\begin{itemize}[leftmargin=1em, nosep]
    \item \textbf{Episodic Memory ($M_E$)}: This component serves as a high-level log of the task, recording key events, major decision points, and sub-task completions. It provides the agent with long-term context regarding the overall task structure and its overarching goals.
    \item \textbf{Working Memory ($M_W$)}: This contains the most recent information, such as the current sub-goal, obstacles encountered, and near-term plans. It is the core component that ensures the continuity of the agent's reasoning across the memory fold.
    \item \textbf{Tool Memory ($M_T$)}: This consolidates all tool-related interactions, including which tools have been used, how they were invoked, and their effectiveness. It allows the agent to learn from its experiences, refining its tool selection and usage strategies.
\end{itemize}

To ensure that the compressed memory is stable and easily parsed by the agent, we employ an \textbf{agent-usable data schema} in JSON format instead of unstructured natural language. It offers two main benefits: maintaining a controllable and predictable structure, and mitigating the loss of critical details that can occur when summarizing long-form text. Details of the data schema are in Appendix~\ref{app:memory_schema}.

\begin{table*}[!t]
\centering
\caption{Main results on general tool usage tasks, encompassing scenarios with both labeled tools and open-set tool retrieval over large-scale toolsets. We report Pass@1 metric for all tasks. For 32B models, the best results are in \textbf{bold} and the second are \underline{underlined}. Results from larger or closed-sourced models are in \textcolor{gray!135}{gray} color for reference.}
\label{tab:tool_usage_performance}
\setlength\tabcolsep{6.8pt}
\fontsize{8.6pt}{10.4pt}\selectfont
\begin{tabular}{lcccccccccccc}
\toprule
\multirow{2}[2]{*}{\textbf{Method}} & \multirow{2}[2]{*}{\textbf{Backbone}}
  & \multicolumn{2}{c}{\textbf{ToolBench}} 
  & \multicolumn{2}{c}{\textbf{API-Bank}} 
  & \multicolumn{2}{c}{\textbf{TMDB}} 
  & \multicolumn{2}{c}{\textbf{Spotify}} 
  & \multicolumn{2}{c}{\textbf{ToolHop}} \\
\cmidrule(lr){3-4} \cmidrule(lr){5-6} \cmidrule(lr){7-8} \cmidrule(lr){9-10} \cmidrule(lr){11-12}
 &  & Success & Path 
   & Success & Path 
   & Success & Path 
   & Success & Path 
   & Correct & Path \\
\midrule
\rowcolor[HTML]{f0f0f0}
\multicolumn{12}{c}{\textit{\textbf{Scenario 1: Completing Tasks w/ Ground-truth Tools}}} \\
\multicolumn{12}{l}{\textit{\textbf{Workflow-based Methods}}} \\
ReAct        & Qwen2.5-32B   & 41.0  & 64.7  & 60.4  & 68.3  & 46.0  & 65.3  & 29.8  & 56.3  & 37.6  & 49.1   \\
CodeAct      & Qwen2.5-32B   & 53.0  & 68.3  & 62.4  & 70.6  & 48.0  & 67.4  & 33.3  & 58.7  & 34.7  & 48.8   \\
Plan-and-Solve   & Qwen2.5-32B   & 52.0  & 65.4  & 58.4  & 67.5  & 51.0  & 71.6  & 28.1  & 54.8  & 39.2  & 49.7   \\
\hdashline
ReAct        & QwQ-32B       & 52.0  & 61.6  & 73.3  & 78.6  & 43.0  & 65.3  & 47.4  & 69.4  & 47.4  & 51.6   \\
CodeAct      & QwQ-32B       & 54.0  & 63.4  & 74.3  & 79.4  & 55.0  & 74.5  & 52.6  & 75.4  & 43.2  & 53.4   \\
Plan-and-Solve   & QwQ-32B       & 55.0  & 64.7  & 70.3  & 75.4  & 48.0  & 61.3  & 49.1  & 70.6  & 45.4  & 50.6   \\
\hdashline
ReAct        & Qwen2.5-72B   & \textcolor{gray!135}{56.0}  & \textcolor{gray!135}{69.3}  & \textcolor{gray!135}{73.3}  & \textcolor{gray!135}{78.6}  & \textcolor{gray!135}{47.0}  & \textcolor{gray!135}{67.7}  & \textcolor{gray!135}{57.9}  & \textcolor{gray!135}{76.6}  & \textcolor{gray!135}{44.8}  & \textcolor{gray!135}{55.4}   \\
ReAct        & GPT-4o        & \textcolor{gray!135}{52.0}  & \textcolor{gray!135}{53.9}  & \textcolor{gray!135}{79.2}  & \textcolor{gray!135}{83.3}  & \textcolor{gray!135}{77.0}  & \textcolor{gray!135}{89.3}  & \textcolor{gray!135}{47.4}  & \textcolor{gray!135}{70.6}  & \textcolor{gray!135}{40.0}  & \textcolor{gray!135}{53.7}   \\
ReAct        & DeepSeek-R1   & \textcolor{gray!135}{57.0}  & \textcolor{gray!135}{68.3}  & \textcolor{gray!135}{71.3}  & \textcolor{gray!135}{76.2}  & \textcolor{gray!135}{76.0}  & \textcolor{gray!135}{89.0}  & \textcolor{gray!135}{64.9}  & \textcolor{gray!135}{81.3}  & \textcolor{gray!135}{50.2}  & \textcolor{gray!135}{61.8}   \\
\multicolumn{12}{l}{\textit{\textbf{Autonomous Tool Usage within Reasoning}}} \\
\rowcolor[HTML]{ecf0ff}
DeepAgent-32B-Base & QwQ-32B     & \underline{63.0} & \underline{74.3} & \textbf{76.2} & \textbf{81.0} & \underline{85.0} & \underline{92.0} & \underline{70.2} & \underline{89.3} & \underline{49.1} & \underline{59.8}  \\
\rowcolor[HTML]{ecf0ff}
DeepAgent-32B-RL   & QwQ-32B     & \textbf{69.0}    & \textbf{78.6} & \underline{75.3} & \underline{80.2}   & \textbf{89.0} & \textbf{94.8}   & \textbf{75.4}   & \textbf{92.0} & \textbf{51.3}   & \textbf{62.5}   \\
\midrule
\rowcolor[HTML]{f0f0f0}
\multicolumn{12}{c}{\textit{\textbf{Scenario 2: Completing Tasks w/ Open-Set Tool Retrieval}}} \\
\multicolumn{12}{l}{\textit{\textbf{Workflow-based Methods}}} \\
ReAct        & Qwen2.5-32B   & 55.0  & 20.8  & 16.0  & 42.0  & 11.0  & 34.5  & 7.0   & 25.4  & 13.2  & 17.9   \\
CodeAct      & Qwen2.5-32B   & 51.0  & 19.0  & \underline{22.0}  & 49.6  & 19.0  & 46.8  & 10.5  & 31.6  & 12.7  & 17.4   \\
Plan-and-Solve   & Qwen2.5-32B   & 54.0  & 20.4  & 18.0  & 42.8  & 15.0  & 40.5  & 8.8   & 26.3  & 12.0  & 16.3   \\
\hdashline
ReAct        & QwQ-32B       & 44.0  & 19.0  & 20.0  & 52.7  & 18.0  & 40.3  & 22.8  & 45.5  & 27.1  & 22.3   \\
CodeAct      & QwQ-32B       & 48.0  & 21.6  & 16.0  & 45.0  & 31.0  & 52.8  & 24.6  & 49.6  & 29.0  & 26.1   \\
Plan-and-Solve   & QwQ-32B       & 45.0  & 19.6  & 18.0  & 44.3  & 24.0  & 46.8  & 19.3  & 42.7  & 25.7  & 20.8   \\
\hdashline
ReAct        & Qwen2.5-72B   & \textcolor{gray!135}{52.0}  & \textcolor{gray!135}{21.6}  & \textcolor{gray!135}{14.0}  & \textcolor{gray!135}{38.9}  & \textcolor{gray!135}{28.0}  & \textcolor{gray!135}{50.7}  & \textcolor{gray!135}{21.1}  & \textcolor{gray!135}{48.5}  & \textcolor{gray!135}{21.1}  & \textcolor{gray!135}{19.9}   \\
ReAct        & GPT-4o        & \textcolor{gray!135}{41.0}  & \textcolor{gray!135}{28.9}  & \textcolor{gray!135}{18.0}  & \textcolor{gray!135}{42.8}  & \textcolor{gray!135}{35.0}  & \textcolor{gray!135}{56.8}  & \textcolor{gray!135}{17.5}  & \textcolor{gray!135}{26.3}  & \textcolor{gray!135}{24.1}  & \textcolor{gray!135}{28.6}   \\
ReAct        & DeepSeek-R1   & \textcolor{gray!135}{47.0}  & \textcolor{gray!135}{22.3}  & \textcolor{gray!135}{12.0}  & \textcolor{gray!135}{57.3}  & \textcolor{gray!135}{34.0}  & \textcolor{gray!135}{53.1}  & \textcolor{gray!135}{29.8}  & \textcolor{gray!135}{51.7}  & \textcolor{gray!135}{36.2}  & \textcolor{gray!135}{32.9}   \\
\multicolumn{12}{l}{\textit{\textbf{Autonomous Tool Retrieval and Usage within Reasoning}}} \\
\rowcolor[HTML]{ecf0ff}
DeepAgent-32B-Base & QwQ-32B     & \underline{60.0} & \underline{35.7} & \underline{22.0} & \underline{61.8} & \underline{52.0} & \underline{71.8} & \underline{49.1} & \underline{68.6} & \underline{38.4} & \underline{40.3}  \\
\rowcolor[HTML]{ecf0ff}
DeepAgent-32B-RL   & QwQ-32B     & \textbf{64.0}       & \textbf{37.2}    & \textbf{24.0} & \textbf{64.9}   & \textbf{55.0} & \textbf{74.3}   & \textbf{50.9}   & \textbf{74.4} & \textbf{40.6}   & \textbf{40.5}   \\
\bottomrule
\end{tabular}
\end{table*}

\begin{table*}[!t]
\centering
\caption{Main results on downstream task applications, spanning Embodied AI (ALFWorld), Online Shopping (WebShop), General AI Assistants (GAIA), and Humanity's Last Exam (HLE). We report Pass@1 for all tasks. For 32B models, the best results are in \textbf{bold} and the second are \underline{underlined}. Results from larger or closed-sourced models are in \textcolor{gray!135}{gray} color for reference. %`$^\dagger$' denotes results from their official releases.
}
\label{tab:application_preformance}
\setlength\tabcolsep{6.9pt}
\fontsize{8.6pt}{10.6pt}\selectfont
\begin{tabular}{lccccccccccccc}
% \begin{tabular}{lc *{11}{>{\centering\arraybackslash}m{6mm}}}
\toprule
\multirow{2}[2]{*}{\textbf{Method}} & \multirow{2}[2]{*}{\textbf{Backbone}} 
  & \multicolumn{2}{c}{\textbf{ALFWorld}} 
  & \multicolumn{2}{c}{\textbf{WebShop}} 
  & \multicolumn{4}{c}{\textbf{GAIA}} 
  & \multicolumn{3}{c}{\textbf{HLE}} \\
  % & \multicolumn{3}{c}{\textbf{Human. Last Exam}} \\
\cmidrule(lr){3-4} \cmidrule(lr){5-6} \cmidrule(lr){7-10} \cmidrule(lr){11-13}
 &  & Success & Path
   & Success & Score 
   & Text & MM & File & All 
   & Text & MM & All \\
\midrule
\rowcolor[HTML]{f0f0f0}
\multicolumn{13}{c}{\textit{\textbf{Completing Tasks w/ Task-specific Toolsets}}} \\
\multicolumn{13}{l}{\textit{\textbf{Workflow-based Methods}}} \\
ReAct        & Qwen2.5-32B   & 60.4  & 79.1  & 6.0   & 28.8  & 25.2  & 16.7  & 13.2  & 21.2  & 6.5  & 7.1  & 6.6 \\
CodeAct      & Qwen2.5-32B   & 65.7  & 83.3  & 12.4  & 34.5  & 28.2  & 20.8  & 18.4  & 24.8  & 7.5  & 8.0  & 7.6 \\
Reflextion   & Qwen2.5-32B   & 66.4  & 86.0  & 9.2   & 31.6  & 29.1  & 20.8  & 18.4  & 25.5  & 5.9  & 5.3  & 5.8 \\
Plan-and-Solve   & Qwen2.5-32B   & 63.4  & 80.4  & 7.6   & 29.3  & 27.2  & 16.7  & 15.8  & 23.0  & 7.2  & 6.2  & 7.0 \\
\hdashline
ReAct        & QwQ-32B       & 82.1  & 87.8  & 17.2  & 45.3  & 35.0  & 8.3   & 36.8  & 31.5  & 13.2 & 8.8  & 12.2 \\
CodeAct      & QwQ-32B       & 78.4  & 86.2  & 18.0  & 46.4  & 38.8  & 20.8  & 31.6  & 34.5  & 14.2 & 8.0  & 12.8 \\
Reflextion   & QwQ-32B       & 85.1  & 88.4  & 21.6  & 50.4  & 37.9  & 20.8  & 36.8  & 35.2  & 11.9 & 7.1  & 10.8 \\
Plan-and-Solve   & QwQ-32B   & 79.1  & 84.7  & 16.0  & 43.8  & 36.9  & 16.7  & 34.2  & 33.3  & 12.9 & 9.7  & 12.2 \\
\hdashline
AgentLM*       & Llama2-70B   & \textcolor{gray!135}{86.0}  & \textcolor{gray!135}{-}     & \textcolor{gray!135}{-}     & \textcolor{gray!135}{64.9}  & \textcolor{gray!135}{-}     & \textcolor{gray!135}{-}     & \textcolor{gray!135}{-}     & \textcolor{gray!135}{-}     & \textcolor{gray!135}{-}    & \textcolor{gray!135}{-}    & \textcolor{gray!135}{-}    \\
ReAct        & Qwen2.5-72B   & \textcolor{gray!135}{86.5}  & \textcolor{gray!135}{86.5}  & \textcolor{gray!135}{22.0}  & \textcolor{gray!135}{44.5}  & \textcolor{gray!135}{32.0}  & \textcolor{gray!135}{20.8}  & \textcolor{gray!135}{31.6}  & \textcolor{gray!135}{30.3}  & \textcolor{gray!135}{9.0}  & \textcolor{gray!135}{8.0}  & \textcolor{gray!135}{8.8} \\
ReAct        & DeepSeek-R1   & \textcolor{gray!135}{79.1}  & \textcolor{gray!135}{85.8}  & \textcolor{gray!135}{19.6}  & \textcolor{gray!135}{49.7}  & \textcolor{gray!135}{43.7}  & \textcolor{gray!135}{29.2}  & \textcolor{gray!135}{39.5}  & \textcolor{gray!135}{40.6}  & \textcolor{gray!135}{14.2} & \textcolor{gray!135}{8.8}  & \textcolor{gray!135}{13.0} \\
ReAct        & GPT-4o        & \textcolor{gray!135}{65.7}  & \textcolor{gray!135}{87.8}  & \textcolor{gray!135}{15.6}  & \textcolor{gray!135}{52.5}  & \textcolor{gray!135}{35.0}  & \textcolor{gray!135}{16.7}  & \textcolor{gray!135}{36.8}  & \textcolor{gray!135}{32.7}  & \textcolor{gray!135}{13.2} & \textcolor{gray!135}{10.6} & \textcolor{gray!135}{12.6} \\
ReAct        & Claude-4      & \textcolor{gray!135}{93.3}  & \textcolor{gray!135}{91.5}  & \textcolor{gray!135}{20.4}  & \textcolor{gray!135}{56.6}  & \textcolor{gray!135}{56.3}  & \textcolor{gray!135}{37.5}  & \textcolor{gray!135}{52.6}  & \textcolor{gray!135}{52.7}  & \textcolor{gray!135}{15.5} & \textcolor{gray!135}{16.8} & \textcolor{gray!135}{15.8} \\
\multicolumn{13}{l}{\textit{\textbf{Autonomous Tool Usage within Reasoning}}} \\
Deep Research        & OpenAI (o3)   & \textcolor{gray!135}{-}  & \textcolor{gray!135}{-}  & \textcolor{gray!135}{-}  & \textcolor{gray!135}{-}  & \textcolor{gray!135}{-}     & \textcolor{gray!135}{-}     & \textcolor{gray!135}{-}     & \textcolor{gray!135}{67.4}  & \textcolor{gray!135}{-}    & \textcolor{gray!135}{-}    & \textcolor{gray!135}{26.6} \\
WebThinker        & QwQ-32B   & -  & -  & -  & -  & 48.5  & 25.0  & 13.2  & 37.0  & 14.2 & 8.8  & 13.0 \\
HiRA        & QwQ-32B   & 84.3  & 87.6  & 23.2  & 51.9  & 44.7  & \underline{33.3}  & 42.1  & 42.5  & 14.5 & 10.6 & 13.6 \\
\hdashline
\rowcolor[HTML]{ecf0ff}
DeepAgent-32B-Base & QwQ-32B     & \underline{88.1} & \underline{91.4} & \underline{32.0} & \underline{55.4} & \underline{49.5} & \textbf{37.5} & \underline{44.7} & \underline{46.7} & \underline{19.1} & \underline{13.3} & \underline{17.8}  \\
\rowcolor[HTML]{ecf0ff}
DeepAgent-32B-RL   & QwQ-32B     & \textbf{91.8}    & \textbf{92.0} & \textbf{34.4}   & \textbf{56.3}   & \textbf{58.3}     & \underline{33.3}     & \textbf{52.6}     & \textbf{53.3}     & \textbf{21.7} & \textbf{15.0} & \textbf{20.2} \\
\bottomrule
\end{tabular}
\end{table*}

\subsection{End-to-end RL Training with ToolPO}
We train DeepAgent end-to-end with Tool Policy Optimization (ToolPO), an RL approach designed for general tool-using agents. %ToolPO builds upon Group Relative Policy Optimization (GRPO), a critic-free variant of PPO that forgoes a value network. Instead, it estimates advantages by normalizing rewards within a group of trajectories sampled from the same prompt, which enhances training stability, especially for large models.

\paragraph{Training Data Collection}
We first collect a diverse training dataset spanning four categories. To instill \textbf{general tool-use} capabilities, we use ToolBench~\cite{ToolLLM}. For \textbf{real-world interaction}, we leverage ALFWorld~\cite{ALFWorld} and WebShop~\cite{WebShop}. To enhance \textbf{deep research} skills, we incorporate data from WebDancer~\cite{WebDancer} and WebShaperQA~\cite{WebShaper}. Lastly, to improve \textbf{mathematical reasoning} with code, we use DeepMath~\cite{DeepMath}. Further details are available in Appendix~\ref{app:training_data_collection}.

\paragraph{Tool Simulator}
Training an agent that interacts with thousands of real-world APIs is often impractical due to instability, latency, and cost. To address this, we develop an \textbf{LLM-based Tool Simulator}. This simulator, powered by an auxiliary LLM, mimics the responses of real-world APIs (e.g., RapidAPI). This approach provides a stable, efficient, and low-cost environment for robust RL training.

\paragraph{Global and Tool-Call Advantage Attribution}
For each input prompt, we sample a group of $K$ trajectories $\{\tau_1, \dots, \tau_K\}$. ToolPO defines two distinct reward components. The first is a reward for overall task success, $R_{\text{succ}}(\tau)$, which is a task-success score that reflects the quality of the final outcome (e.g., the accuracy of the final answer). The second is a tool-call reward, $R_{\text{action}}(\tau)$, which reflects the quality of intermediate actions. This action-level reward is composed of rewards for correct tool invocations and efficient memory folding. Specifically, $R_{\text{action}}(\tau) = \lambda_1 \sum_{t=1}^{T} C(a_t^{\text{call}}) + \lambda_2 S_{\text{pref}}(\tau)$, where $C(a_t^{\text{call}})$ is 1 if a tool call is correct and 0 otherwise. $S_{\text{pref}}(\tau)$ is a preference score encouraging efficient use of memory folding, defined by comparing a trajectory with folding ($\tau_{\text{fold}}$) to one without ($\tau_{\text{direct}}$): $S_{\text{pref}} = (L(\tau_{\text{direct}}) - L(\tau_{\text{fold}})) / (L(\tau_{\text{direct}}) + L(\tau_{\text{fold}}))$.

Based on these rewards, we compute two separate group-relative advantages. The task success advantage for trajectory $\tau_k$ is:
\begin{equation}
    A_{\text{succ}}(\tau_k) = R_{\text{succ}}(\tau_k) - \frac{1}{K} \sum\nolimits_{j=1}^{K} R_{\text{succ}}(\tau_j),
\end{equation}
which is attributed to all generated tokens in $\tau_k$, providing a global learning signal. Similarly, the action-level advantage is:
\begin{equation}
    A_{\text{action}}(\tau_k) = R_{\text{action}}(\tau_k) - \frac{1}{K} \sum\nolimits_{j=1}^{K} R_{\text{action}}(\tau_j).
\end{equation}
Crucially, this advantage is attributed \textit{only} to the specific tokens that constitute the tool call and memory folding actions. This fine-grained credit assignment provides a more targeted signal for learning correct and efficient tool use.

\paragraph{Optimization Objective}
The total advantage for a given token $y_i$ in trajectory $\tau_k$ is the sum of the global and local advantages:
\begin{equation}
    A(y_i) = A_{\text{succ}}(\tau_k) + M(y_i) \cdot A_{\text{action}}(\tau_k),
\end{equation}
where $M(y_i)$ is a mask that is 1 if $y_i$ is part of a tool-call or memory-fold token sequence, and 0 otherwise. ToolPO then optimizes the policy using a clipped surrogate objective function:
\begin{equation}
\begin{aligned}
    &\mathcal{L}_{\text{ToolPO}}(\theta) = \\
    &\mathbb{E}_{\tau_k}\left[ \sum\nolimits_{i=1}^{|\tau_k|} \min\Big( \rho_i(\theta) A(y_i), \text{clip}(\rho_i(\theta), 1 - \epsilon, 1 + \epsilon) A(y_i) \Big) \right].
\end{aligned}
\end{equation}
Here, $\rho_i(\theta) = \frac{\pi_\theta(y_i|y_{<i}, s)}{\pi_{\theta_{\text{old}}}(y_i|y_{<i}, s)}$ is the probability ratio for token $y_i$. This objective encourages the model to increase the probability of both intermediate actions and end-to-end task accomplishment that exhibit positive relative advantage, thereby ensuring stable and effective policy updates.

\section{Experimental Settings}
\subsection{Tasks and Datasets}
We conduct extensive experiments on a wide range of benchmarks, including general tool-use and downstream applications.

\textit{General Tool-Use.}
These benchmarks cover toolsets from tens to \(>10\)k tools, and thus stress scalability. They evaluate core capabilities for general tool use, including tool planning, tool retrieval, and accurate multi-step tool calling. We use \textbf{ToolBench~\cite{ToolLLM}} (16k+ real-world APIs; G3 subset with multi-step/multi-tool calls), \textbf{API-Bank~\cite{API-Bank}} (314 dialogues; 73 APIs; 753 calls), \textbf{RestBench~\cite{RestGPT}} (TMDB: 54 tools, 2.3 calls/question; Spotify: 40 tools, 2.6 calls/question), and \textbf{ToolHop~\cite{ToolHop}} (3,912 executable tools; 3--7 calls/task). We evaluate two settings: provided ground-truth tools and open-set tool retrieval from the full toolset.

\textit{Downstream Applications.}
We also evaluate downstream applications with domain-specific toolsets: \textbf{ALFWorld~\cite{ALFWorld}} (text embodied tasks with nine actions, e.g., move/take), \textbf{WebShop~\cite{WebShop}} (shopping with `search' and `click'), \textbf{GAIA~\cite{GAIA}} (web search/browsing, VQA, code, file reading), and \textbf{Humanity's Last Exam (HLE)~\cite{HLE}} (code, search, browsing, VQA). These tasks test long-horizon interaction in more realistic environments, requiring state tracking, error recovery, and coordination across heterogeneous tools; we equip agents with task-specific toolsets.
\subsection{Baselines}
Our baselines include: (1) \textbf{Workflow-based Methods}: ReAct~\cite{yao2022react} alternates explicit reasoning with environment actions in a Reason-Act-Observe loop. CodeAct~\cite{CodeAct} expresses actions as executable Python code that runs in an interpreter. Plan-and-Solve~\cite{Plan-and-Solve} first sketches a high-level plan and then executes it step by step. Reflexion~\cite{Reflexion} enhances learning through verbal self-reflection after failed attempts. AgentLM~\cite{AgentTuning} uses instruction tuning to enhance general agent capabilities of LLMs. (2) \textbf{Autonomous Tool Usage within Reasoning}: WebThinker~\cite{WebThinker} interleaves thinking with web search and deep web exploration. HiRA~\cite{HiRA} introduces a hierarchical agent architecture where a meta planner decomposes tasks, a coordinator routes subtasks, and specialized executors solve them with dual-channel memory. OpenAI Deep Research~\cite{openai_deep_research} is an agentic system based on reasoning models.
%
%
%\vspace*{-0.7mm}
\subsection{Implementation Details}
We use QwQ-32B~\cite{qwen_qwq} as DeepAgent's backbone model, with Qwen2.5-32B-Instruct~\cite{qwen2.5} as the auxiliary model in our main results. Text generation employs a maximum of 81,920 tokens with temperature 0.7, top\_p 0.8, top\_k 20, and repetition penalty 1.05. Web search and page browsing are implemented using Google Serper API and Jina Reader API, respectively. The VQA tool is based on Qwen2.5-VL-32B-Instruct~\cite{Qwen2.5-VL}. Tool retrieval is performed using bge-large-en-v1.5~\cite{C-Pack}. Training consists of 100 steps of ToolPO with batch size 64, $\lambda_1=\lambda_2=1$, rollout size $K=8$, and maximum sequence length 32,768. Additional details are provided in Appendix~\ref{app:implementation_details}. All experiments are conducted on 64 NVIDIA H20-141GB GPUs.

\begin{figure}[!t]
\centering
\includegraphics[width=1\linewidth]{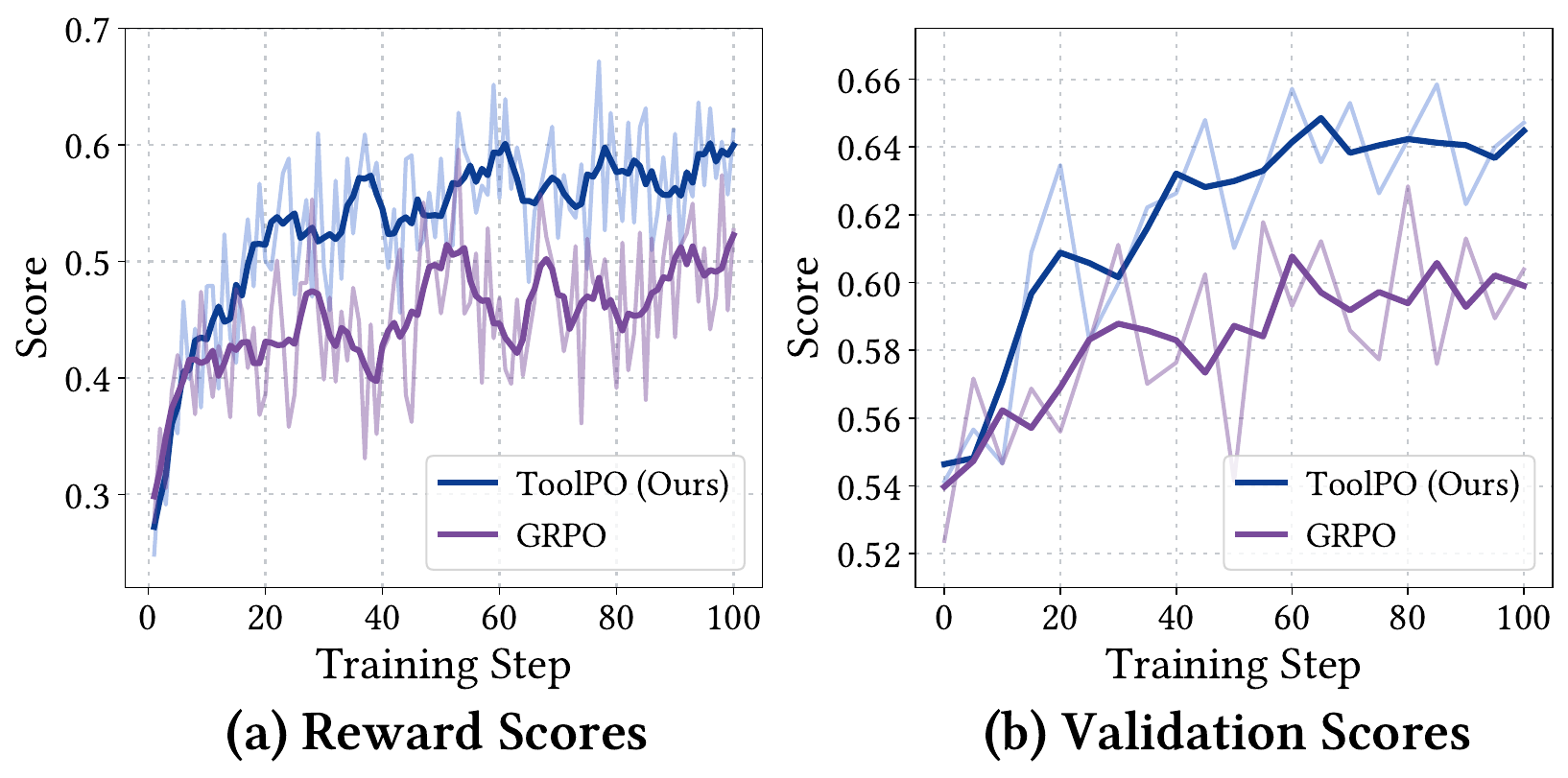}
\caption{Visualization of training dynamics, including (a) reward scores and (b) validation scores across training steps.}
\label{fig:training_curve}
\end{figure}

\begin{table}[!t]
\centering
\caption{Ablation studies on the components of DeepAgent, where the best results are in \textbf{bold}.}
\label{tab:ablation}
\setlength\tabcolsep{2.4pt}
\begin{tabular}{lccccc}
\toprule
\multirow{2}[2]{*}{\textbf{Method}} & \multicolumn{2}{c}{\textbf{Tool-Usage}} & \multicolumn{2}{c}{\textbf{Application}} & \multirow{2}{*}{\textbf{ Avg. }} \\
\cmidrule(lr){2-3} \cmidrule(lr){4-5}
 & ToolB. & ToolH. & WebS. & GAIA & \\
\midrule
\rowcolor[HTML]{ecf0ff}
DeepAgent-32B-RL & \textbf{64.0} & \textbf{40.6} & \textbf{34.4} & \textbf{53.3} & \textbf{48.1} \\
\midrule
~~w/o Training (Base) & 60.0 & 38.4 & 32.0 & 46.7 & 44.3 \\
~~w/o Memory Folding & 63.0 & 36.6 & 32.4 & 44.7 & 44.2 \\
~~w/o Tool Simulation & 62.0 & 35.2 & 33.6 & 48.5 & 44.8 \\
~~w/o Tool Adv. Attribution & 62.0 & 39.6 & 33.2 & 49.5 & 46.1 \\
\bottomrule
\end{tabular}
\end{table}

\section{Experimental Results}
\subsection{Main Results on General Tool Usage Tasks}
Table~\ref{tab:tool_usage_performance} summarizes results on general tool-use tasks and yields three observations.
\textbf{(1) DeepAgent's End-to-End Reasoning Surpasses Workflow-Based Methods.}
DeepAgent consistently outperforms workflow-based agents. On labeled-tool tasks, DeepAgent-32B-RL reaches 89.0\% on TMDB and 75.4\% on Spotify, exceeding the best 32B baselines (55.0\% and 52.6\%). This highlights the advantage of end-to-end agentic reasoning over rigid, predefined action loops.
\textbf{(2) DeepAgent Maintains Robustness in Open-Set Scenarios.}
Gains are larger in open-set settings where tool discovery is required: on ToolBench and ToolHop, DeepAgent-32B-RL achieves 64.0\% and 40.6\%, surpassing the best baselines (54.0\% and 29.0\%). This suggests that on-demand tool discovery within the reasoning process is both more robust and more scalable in realistic open-set tool environments.
\textbf{(3) ToolPO Training Further Improves Tool-Usage Capabilities.}
ToolPO yields consistent improvements over the base model, increasing ToolBench success by up to 6.0\% and Spotify (labeled) by 5.2\%. These gains indicate that our RL training better aligns intermediate tool calls with end-task success.

\subsection{Main Results on Downstream Applications}
Table~\ref{tab:application_preformance} reports the downstream results that require long-horizon interaction and more complex environment dynamics.
\textbf{(1) The autonomous reasoning paradigm generally outperforms the workflow-based methods.} 
Methods that integrate tool use into continuous reasoning outperform workflow-based agents. On GAIA, DeepAgent-32B-Base (46.7) and HiRA (42.5) exceed the best workflow baseline CodeAct (34.5); on WebShop, DeepAgent-32B-Base scores 32.0 vs. 18.0. This supports that long-horizon tasks benefit from flexible, integrated reasoning-and-action rather than fixed workflows.
\textbf{(2) DeepAgent demonstrates superior performance across various application tasks.} 
DeepAgent achieves the best performance among 32B models: 53.3 on GAIA (vs. 42.5 for HiRA) and 91.8\% on ALFWorld (vs. 84.3). We attribute this to DeepAgent's coherent reasoning process and its support for robust long-horizon interaction (e.g., autonomous memory folding).
\textbf{(3) ToolPO training further improves performance on downstream applications.} 
ToolPO further improves downstream performance: GAIA 46.7 $\rightarrow$ 53.3 (+6.6) and ALFWorld 88.1\% $\rightarrow$ 91.8\% (+3.7). This shows the tool-use improvements learned by ToolPO transfer to interactive downstream settings.

\begin{table}[!t]
\centering
\caption{Effectiveness analysis of autonomous tool retrieval strategy in open-set scenarios compared to pre-retrieved tool methods. Numbers in parentheses indicate toolset sizes.}
\label{tab:tool_retrieval}
\setlength\tabcolsep{3pt}
\begin{tabular}{p{2.7cm}ccccc}
\toprule
\multirow{2}{*}{\textbf{Method}} & \textbf{ToolB.} & \textbf{ToolH.} & \textbf{TMDB} & \textbf{Spotify} & \multirow{2}{*}{\textbf{Avg.}} \\
 & \textbf{(16k)} & \textbf{(3.9k)} & \textbf{(54)} & \textbf{(40)} & \\
\midrule
\multicolumn{6}{l}{\textbf{\textit{ReAct Workflow}}} \\
Input Retrieved Tool & 35.0 & 25.4 & 14.0 & 15.0 & 22.4 \\
Auto. Tool Retrieval & 34.0 & 37.1 & 18.0 & 27.8 & 28.0 \\
\midrule
\multicolumn{6}{l}{\textbf{\textit{Plan-and-Solve Workflow}}} \\
Input Retrieved Tool & 37.0 & 24.8 & 19.0 & 16.0 & 24.2 \\
Auto. Tool Retrieval & 45.0 & 25.7 & 24.0 & 19.3 & 28.5 \\
\midrule
\multicolumn{6}{l}{\textbf{\textit{End-to-end Agentic Reasoning (DeepAgent)}}} \\
Input Retrieved Tool & 53.0 & 37.0 & 34.0 & 43.9 & 42.0 \\
\rowcolor[HTML]{ecf0ff}
Auto. Tool Retrieval & \textbf{64.0} & \textbf{40.6} & \textbf{55.0} & \textbf{50.9} & \textbf{52.6} \\
\bottomrule
\end{tabular}
\end{table}

\subsection{Analysis of Training Dynamics}
Figure~\ref{fig:training_curve} shows the training dynamics of DeepAgent, including the reward scores and validation scores across training steps.
As shown in the figure, \textbf{(1) DeepAgent trained with ToolPO achieves higher upper bounds on both reward and validation scores compared to the commonly used GRPO.} \textbf{(2) Moreover, the training reward exhibits less fluctuation than GRPO, demonstrating better training stability.} This indicates that using tool simulators instead of directly training with unstable real-world APIs, along with employing tool-call process supervision, enables more stable and effective training of tool-usage capabilities.

\begin{figure}[!t]
\centering
\includegraphics[width=1\linewidth]{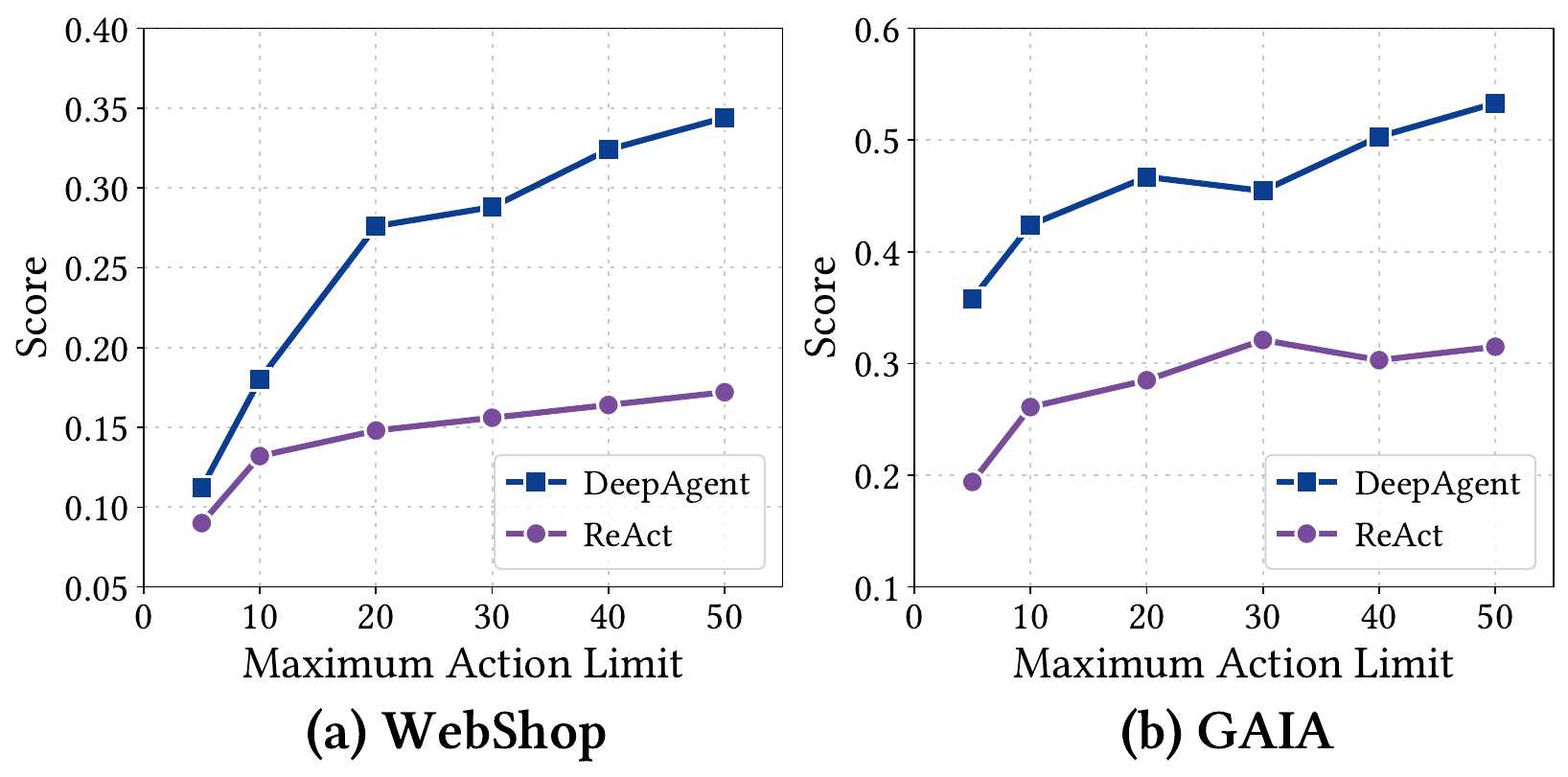}
\caption{Scaling analysis of performance with respect to maximum action limits on WebShop and GAIA datasets.}
\label{fig:action_scaling}
\end{figure}

\subsection{Ablation Studies}
We conduct ablation studies in Table~\ref{tab:ablation} to validate the effectiveness of each component in DeepAgent.
\textbf{(1) Importance of ToolPO Training:} Removing ToolPO training (the Base model) results in the most significant performance drop (from 48.1 to 44.3). This highlights the central role of our end-to-end RL method in enhancing tool use and complex task completion.
\textbf{(2) Effectiveness of Memory Folding:} The absence of memory folding also leads to a substantial performance decline (average score drops to 44.2), particularly on the long-horizon task GAIA (from 53.3 to 44.7). This confirms that the autonomous memory folding mechanism, allowing the agent to "take a breath" and replan, is crucial for robust long-term interaction.
\textbf{(3) Contribution of Training Strategies:} Removing the tool simulator and tool-call advantage attribution both lead to performance degradation. This validates that the tool simulator enables more stable training, and fine-grained advantage attribution provides precise learning signals.

\subsection{Effectiveness of Tool Retrieval Strategies}
To compare pre-retrieving tools versus autonomous discovery during task execution, we conduct experiments shown in Table~\ref{tab:tool_retrieval}. We find:
\textbf{(1) The on-demand nature of dynamic tool discovery yields superior performance and robust scalability.} 
 Autonomous tool retrieval during reasoning consistently outperforms pre-retrieved tools across all frameworks, demonstrating the superiority of on-demand tool access in open-set scenarios. Performance gains are most pronounced on large toolsets like ToolBench (16k tools) and ToolHop (3.9k tools), indicating robust scalability for real-world tasks.
\textbf{(2) DeepAgent synergizes better with dynamic retrieval.}
Combined with autonomous tool retrieval, our framework achieves the best results by a large margin, scoring 52.6 on average versus 28.5 for the best workflow-based method. This demonstrates that DeepAgent's architecture is uniquely suited for dynamic tool discovery.

\subsection{Scaling Analysis of Action Limits}
Figure~\ref{fig:action_scaling} illustrates the performance of DeepAgent and ReAct on the WebShop and GAIA datasets as the maximum action limit is varied. The results yield several key insights. 
\textbf{(1) DeepAgent consistently and significantly outperforms the ReAct baseline across all tested action limits on both datasets}, demonstrating its superior effectiveness. 
\textbf{(2) For both agents, performance generally improves as the maximum number of actions increases.} This suggests that complex tasks benefit from a longer interaction horizon, allowing for more thorough exploration and reasoning.
\textbf{(3) DeepAgent exhibits stronger scalability.} As the action limit increases, the performance gap between DeepAgent and ReAct widens, particularly on WebShop. This sustained gain suggests DeepAgent strategically selects effective, task-relevant actions, avoiding the wasteful steps that limit ReAct's scalability.
% \gl{
% This sustained gain suggests DeepAgent strategically selects effective, task-relevant actions, avoiding the wasteful steps that limit 
% ReAct's scalability.
% }

\begin{table}[!t]
\centering
\caption{Performance with different reasoning model backbones: MOE-based models with 30B and 235B parameters.}
\label{tab:backbone_comparison}
\setlength\tabcolsep{3.2pt}
\begin{tabular}{lcccccc}
\toprule
\multirow{2}[2]{*}{\textbf{Method}} & \multicolumn{2}{c}{\textbf{Tool-Usage}} & \multicolumn{3}{c}{\textbf{Application}} & \multirow{2}{*}{\textbf{Avg.}} \\
\cmidrule(lr){2-3} \cmidrule(lr){4-6}
 & ToolB. & ToolH. & ALF. & WebS. & GAIA & \\
\midrule
\multicolumn{7}{l}{\textbf{\textit{Qwen3-30B-A3B-Thinking}}} \\
ReAct & 52.0 & 22.0 & 67.9 & 18.4 & 34.5 & 35.7 \\
Plan-and-Solve & 50.0 & 23.6 & 68.7 & 20.4 & 35.2 & 37.0 \\
\rowcolor[HTML]{ecf0ff}
DeepAgent (Base) & \textbf{59.0} & \textbf{47.5} & \textbf{69.4} & \textbf{31.4} & \textbf{39.4} & \textbf{46.9} \\
\midrule
\multicolumn{7}{l}{\textbf{\textit{Qwen3-235B-A22B-Thinking}}} \\
ReAct & 61.0 & 40.9 & 79.9 & 21.6 & 36.4 & 45.1 \\
Plan-and-Solve & 63.0 & 43.0 & 78.4 & 24.4 & 38.4 & 46.0 \\
\rowcolor[HTML]{ecf0ff}
DeepAgent (Base) & \textbf{67.0} & \textbf{48.2} & \textbf{85.8} & \textbf{37.2} & \textbf{51.5} & \textbf{55.7} \\
\bottomrule
\end{tabular}
\end{table}

% \subsection{Effectiveness of DeepAgent with Different Backbone LRMs}
\subsection{Generalization Across Different Backbones}
Table~\ref{tab:backbone_comparison} shows the performance of DeepAgent with different backbone large reasoning models, including Qwen3-30B-A3B-Thinking and Qwen3-235B-A22B-Thinking~\cite{Qwen3}.
\textbf{(1) DeepAgent consistently outperforms workflow-based methods.} With both the 30B and 235B MoE-based reasoning models as backbones, DeepAgent maintains a significant performance margin over ReAct and Plan-and-Solve, demonstrating the generalizability of its agentic reasoning approach.
\textbf{(2) DeepAgent scales effectively with larger models.} While all methods benefit from scaling the backbone from a 30B to a 235B model, DeepAgent shows the largest absolute performance gains on complex application tasks.
% \textbf{(3) MoE architectures show a trade-off between efficiency and capability.} With a similar total parameter, the MoE-based `Qwen3-30B-A3B-Thinking` (with only 3B activated parameters) exhibits slightly lower performance compared to the dense `QwQ-32B` model (as shown in Tables~\ref{tab:tool_usage_performance}--\ref{tab:application_preformance}), particularly on complex application benchmarks.
% \gl{Omit point (3) if the conclusion cannot be clearly substantiated.}

\section{Conclusion}
\label{sec:conclusion}
In this work, we introduce DeepAgent, an end-to-end reasoning agent that unifies thinking, tool discovery, and execution into a single, coherent agentic reasoning process. To enable robust long-horizon interaction, we propose an autonomous memory folding mechanism that compresses interaction history into a structured memory, allowing the agent to "take a breath" and reconsider its strategy. We also introduce ToolPO, an end-to-end RL method that leverages LLM simulated APIs for stable training and fine-grained advantage attribution for precise credit assignment to tool invocations. Extensive experiments on general tool-use and downstream applications demonstrate that DeepAgent significantly outperforms various baseline agents, particularly in open-set scenarios requiring dynamic tool discovery over scalable toolsets. 
This work opens new avenues for developing more general and scalable LLM agents for broader real-world applications.
% This work opens new avenues for creating more general and capable agents, with future research directions including more advanced memory management, and expansion into multimodal and human-collaborative domains.

\begin{acks}
This work was supported by the National Natural Science Foundation of China No. 62272467, and the China Postdoctoral Science Foundation under Grant Number 2025T180440. The work was partially done at the Engineering Research Center of Next-Generation Intelligent Search and Recommendation, MOE.

\end{acks}

\newpage
\balance
\bibliographystyle{ACM-Reference-Format}
\bibliography{refer}

@misc{openai_deep_research,
  author = {OpenAI},
  title = {Introducing deep research},
  year = {2025},
  howpublished = {\url{https://openai.com/index/introducing-deep-research}}
}

@article{qwen_qwq,
  title={Qwq: Reflect deeply on the boundaries of the unknown},
  author={Team, Qwen},
  journal={Hugging Face},
  year={2024}
}

@article{deepseek-r1,
  author       = {DeepSeek{-}AI and
                  Daya Guo and
                  Dejian Yang and
                  Haowei Zhang and
                  Junxiao Song and
                  Ruoyu Zhang and
                  Runxin Xu and
                  Qihao Zhu and
                  Shirong Ma and
                  Peiyi Wang and
                  Xiao Bi and
                  Xiaokang Zhang and
                  Xingkai Yu and
                  Yu Wu and
                  Z. F. Wu and
                  Zhibin Gou and
                  Zhihong Shao and
                  Zhuoshu Li and
                  Ziyi Gao and
                  Aixin Liu and
                  Bing Xue and
                  Bingxuan Wang and
                  Bochao Wu and
                  Bei Feng and
                  Chengda Lu and
                  Chenggang Zhao and
                  Chengqi Deng and
                  Chenyu Zhang and
                  Chong Ruan and
                  Damai Dai and
                  Deli Chen and
                  Dongjie Ji and
                  Erhang Li and
                  Fangyun Lin and
                  Fucong Dai and
                  Fuli Luo and
                  Guangbo Hao and
                  Guanting Chen and
                  Guowei Li and
                  H. Zhang and
                  Han Bao and
                  Hanwei Xu and
                  Haocheng Wang and
                  Honghui Ding and
                  Huajian Xin and
                  Huazuo Gao and
                  Hui Qu and
                  Hui Li and
                  Jianzhong Guo and
                  Jiashi Li and
                  Jiawei Wang and
                  Jingchang Chen and
                  Jingyang Yuan and
                  Junjie Qiu and
                  Junlong Li and
                  J. L. Cai and
                  Jiaqi Ni and
                  Jian Liang and
                  Jin Chen and
                  Kai Dong and
                  Kai Hu and
                  Kaige Gao and
                  Kang Guan and
                  Kexin Huang and
                  Kuai Yu and
                  Lean Wang and
                  Lecong Zhang and
                  Liang Zhao and
                  Litong Wang and
                  Liyue Zhang and
                  Lei Xu and
                  Leyi Xia and
                  Mingchuan Zhang and
                  Minghua Zhang and
                  Minghui Tang and
                  Meng Li and
                  Miaojun Wang and
                  Mingming Li and
                  Ning Tian and
                  Panpan Huang and
                  Peng Zhang and
                  Qiancheng Wang and
                  Qinyu Chen and
                  Qiushi Du and
                  Ruiqi Ge and
                  Ruisong Zhang and
                  Ruizhe Pan and
                  Runji Wang and
                  R. J. Chen and
                  R. L. Jin and
                  Ruyi Chen and
                  Shanghao Lu and
                  Shangyan Zhou and
                  Shanhuang Chen and
                  Shengfeng Ye and
                  Shiyu Wang and
                  Shuiping Yu and
                  Shunfeng Zhou and
                  Shuting Pan and
                  S. S. Li},
  title        = {DeepSeek-R1: Incentivizing Reasoning Capability in LLMs via Reinforcement
                  Learning},
  journal      = {CoRR},
  volume       = {abs/2501.12948},
  year         = {2025},
  url          = {https://doi.org/10.48550/arXiv.2501.12948},
  doi          = {10.48550/ARXIV.2501.12948},
  eprinttype    = {arXiv},
  eprint       = {2501.12948},
  bibsource    = {dblp computer science bibliography, https://dblp.org}
}

@article{2410_o1_journey_part1,
  title={O1 Replication Journey: A Strategic Progress Report--Part 1},
  author={Qin, Yiwei and Li, Xuefeng and Zou, Haoyang and Liu, Yixiu and Xia, Shijie and Huang, Zhen and Ye, Yixin and Yuan, Weizhe and Liu, Hector and Li, Yuanzhi and others},
  journal={arXiv preprint arXiv:2410.18982},
  year={2024}
}

@article{shinn2024reflexion,
  title={Reflexion: Language agents with verbal reinforcement learning},
  author={Shinn, Noah and Cassano, Federico and Gopinath, Ashwin and Narasimhan, Karthik and Yao, Shunyu},
  journal={Advances in Neural Information Processing Systems},
  volume={36},
  year={2024}
}

@article{xue2025simpletir,
  title={Simpletir: End-to-end reinforcement learning for multi-turn tool-integrated reasoning},
  author={Xue, Zhenghai and Zheng, Longtao and Liu, Qian and Li, Yingru and Zheng, Xiaosen and Ma, Zejun and An, Bo},
  journal={arXiv preprint arXiv:2509.02479},
  year={2025}
}

@article{yao2022react,
  title={React: Synergizing reasoning and acting in language models},
  author={Yao, Shunyu and Zhao, Jeffrey and Yu, Dian and Du, Nan and Shafran, Izhak and Narasimhan, Karthik and Cao, Yuan},
  journal={arXiv preprint arXiv:2210.03629},
  year={2022}
}

@inproceedings{unigen,
  author       = {Xiaoxi Li and
                  Yujia Zhou and
                  Zhicheng Dou},
  editor       = {Michael J. Wooldridge and
                  Jennifer G. Dy and
                  Sriraam Natarajan},
  title        = {UniGen: {A} Unified Generative Framework for Retrieval and Question
                  Answering with Large Language Models},
  booktitle    = {Thirty-Eighth {AAAI} Conference on Artificial Intelligence, {AAAI}
                  2024, Thirty-Sixth Conference on Innovative Applications of Artificial
                  Intelligence, {IAAI} 2024, Fourteenth Symposium on Educational Advances
                  in Artificial Intelligence, {EAAI} 2014, February 20-27, 2024, Vancouver,
                  Canada},
  pages        = {8688--8696},
  publisher    = {{AAAI} Press},
  year         = {2024},
  url          = {https://doi.org/10.1609/aaai.v38i8.28714},
  doi          = {10.1609/AAAI.V38I8.28714},
  bibsource    = {dblp computer science bibliography, https://dblp.org}
}

@article{openai2024openaio1card,
  title={OpenAI o1 System Card},
  author={Jaech, Aaron and Kalai, Adam and Lerer, Adam and Richardson, Adam and El-Kishky, Ahmed and Low, Aiden and Helyar, Alec and Madry, Aleksander and Beutel, Alex and Carney, Alex and others},
  journal={arXiv preprint arXiv:2412.16720},
  year={2024}
}

@article{Qwen2.5-VL,
  author       = {Shuai Bai and
                  Keqin Chen and
                  Xuejing Liu and
                  Jialin Wang and
                  Wenbin Ge and
                  Sibo Song and
                  Kai Dang and
                  Peng Wang and
                  Shijie Wang and
                  Jun Tang and
                  Humen Zhong and
                  Yuanzhi Zhu and
                  Ming{-}Hsuan Yang and
                  Zhaohai Li and
                  Jianqiang Wan and
                  Pengfei Wang and
                  Wei Ding and
                  Zheren Fu and
                  Yiheng Xu and
                  Jiabo Ye and
                  Xi Zhang and
                  Tianbao Xie and
                  Zesen Cheng and
                  Hang Zhang and
                  Zhibo Yang and
                  Haiyang Xu and
                  Junyang Lin},
  title        = {Qwen2.5-VL Technical Report},
  journal      = {CoRR},
  volume       = {abs/2502.13923},
  year         = {2025},
  url          = {https://doi.org/10.48550/arXiv.2502.13923},
  doi          = {10.48550/ARXIV.2502.13923},
  eprinttype    = {arXiv},
  eprint       = {2502.13923},
  bibsource    = {dblp computer science bibliography, https://dblp.org}
}

@misc{qwen2.5,
      title={Qwen2.5 Technical Report}, 
      author={Qwen and : and An Yang and Baosong Yang and Beichen Zhang and Binyuan Hui and Bo Zheng and Bowen Yu and Chengyuan Li and Dayiheng Liu and Fei Huang and Haoran Wei and Huan Lin and Jian Yang and Jianhong Tu and Jianwei Zhang and Jianxin Yang and Jiaxi Yang and Jingren Zhou and Junyang Lin and Kai Dang and Keming Lu and Keqin Bao and Kexin Yang and Le Yu and Mei Li and Mingfeng Xue and Pei Zhang and Qin Zhu and Rui Men and Runji Lin and Tianhao Li and Tingyu Xia and Xingzhang Ren and Xuancheng Ren and Yang Fan and Yang Su and Yichang Zhang and Yu Wan and Yuqiong Liu and Zeyu Cui and Zhenru Zhang and Zihan Qiu},
      year={2024},
      eprint={2412.15115},
      archivePrefix={arXiv},
      primaryClass={cs.CL},
      url={https://arxiv.org/abs/2412.15115}, 
}

@article{2501_search_o1,
  author       = {Xiaoxi Li and
                  Guanting Dong and
                  Jiajie Jin and
                  Yuyao Zhang and
                  Yujia Zhou and
                  Yutao Zhu and
                  Peitian Zhang and
                  Zhicheng Dou},
  title        = {Search-o1: Agentic Search-Enhanced Large Reasoning Models},
  journal      = {CoRR},
  volume       = {abs/2501.05366},
  year         = {2025},
  url          = {https://doi.org/10.48550/arXiv.2501.05366},
  doi          = {10.48550/ARXIV.2501.05366},
  eprinttype    = {arXiv},
  eprint       = {2501.05366},
  bibsource    = {dblp computer science bibliography, https://dblp.org}
}

@article{2503_Search_R1,
  author       = {Bowen Jin and
                  Hansi Zeng and
                  Zhenrui Yue and
                  Dong Wang and
                  Hamed Zamani and
                  Jiawei Han},
  title        = {Search-R1: Training LLMs to Reason and Leverage Search Engines with
                  Reinforcement Learning},
  journal      = {CoRR},
  volume       = {abs/2503.09516},
  year         = {2025},
  url          = {https://doi.org/10.48550/arXiv.2503.09516},
  doi          = {10.48550/ARXIV.2503.09516},
  eprinttype    = {arXiv},
  eprint       = {2503.09516},
  bibsource    = {dblp computer science bibliography, https://dblp.org}
}

@article{2502_survey_from_system1,
  author       = {Zhong{-}Zhi Li and
                  Duzhen Zhang and
                  Ming{-}Liang Zhang and
                  Jiaxin Zhang and
                  Zengyan Liu and
                  Yuxuan Yao and
                  Haotian Xu and
                  Junhao Zheng and
                  Pei{-}Jie Wang and
                  Xiuyi Chen and
                  Yingying Zhang and
                  Fei Yin and
                  Jiahua Dong and
                  Zhijiang Guo and
                  Le Song and
                  Cheng{-}Lin Liu},
  title        = {From System 1 to System 2: {A} Survey of Reasoning Large Language
                  Models},
  journal      = {CoRR},
  volume       = {abs/2502.17419},
  year         = {2025},
  url          = {https://doi.org/10.48550/arXiv.2502.17419},
  doi          = {10.48550/ARXIV.2502.17419},
  eprinttype    = {arXiv},
  eprint       = {2502.17419},
  bibsource    = {dblp computer science bibliography, https://dblp.org}
}

@article{2503_Towards_Reasoning_Era,
  author       = {Qiguang Chen and
                  Libo Qin and
                  Jinhao Liu and
                  Dengyun Peng and
                  Jiannan Guan and
                  Peng Wang and
                  Mengkang Hu and
                  Yuhang Zhou and
                  Te Gao and
                  Wanxiang Che},
  title        = {Towards Reasoning Era: {A} Survey of Long Chain-of-Thought for Reasoning
                  Large Language Models},
  journal      = {CoRR},
  volume       = {abs/2503.09567},
  year         = {2025},
  url          = {https://doi.org/10.48550/arXiv.2503.09567},
  doi          = {10.48550/ARXIV.2503.09567},
  eprinttype    = {arXiv},
  eprint       = {2503.09567},
  bibsource    = {dblp computer science bibliography, https://dblp.org}
}

@article{2504_DeepResearcher,
  title={DeepResearcher: Scaling Deep Research via Reinforcement Learning in Real-world Environments},
  author={Zheng, Yuxiang and Fu, Dayuan and Hu, Xiangkun and Cai, Xiaojie and Ye, Lyumanshan and Lu, Pengrui and Liu, Pengfei},
  journal={arXiv preprint arXiv:2504.03160},
  year={2025}
}

@inproceedings{GAIA,
  author       = {Gr{\'{e}}goire Mialon and
                  Cl{\'{e}}mentine Fourrier and
                  Thomas Wolf and
                  Yann LeCun and
                  Thomas Scialom},
  title        = {{GAIA:} a benchmark for General {AI} Assistants},
  booktitle    = {The Twelfth International Conference on Learning Representations,
                  {ICLR} 2024, Vienna, Austria, May 7-11, 2024},
  publisher    = {OpenReview.net},
  year         = {2024},
  url          = {https://openreview.net/forum?id=fibxvahvs3},
  bibsource    = {dblp computer science bibliography, https://dblp.org}
}

@article{HLE,
  author       = {Long Phan and
                  Alice Gatti and
                  Ziwen Han and
                  Nathaniel Li and
                  Josephina Hu and
                  Hugh Zhang and
                  Sean Shi and
                  Michael Choi and
                  Anish Agrawal and
                  Arnav Chopra and
                  Adam Khoja and
                  Ryan Kim and
                  Jason Hausenloy and
                  Oliver Zhang and
                  Mantas Mazeika and
                  Daron Anderson and
                  Tung Nguyen and
                  Mobeen Mahmood and
                  Fiona Feng and
                  Steven Y. Feng and
                  Haoran Zhao and
                  Michael Yu and
                  Varun Gangal and
                  Chelsea Zou and
                  Zihan Wang and
                  Jessica P. Wang and
                  Pawan Kumar and
                  Oleksandr Pokutnyi and
                  Robert Gerbicz and
                  Serguei Popov and
                  John{-}Clark Levin and
                  Mstyslav Kazakov and
                  Johannes Schmitt and
                  Geoff Galgon and
                  Alvaro Sanchez and
                  Yongki Lee and
                  Will Yeadon and
                  Scott Sauers and
                  Marc Roth and
                  Chidozie Agu and
                  S{\o}ren Riis and
                  Fabian Giska and
                  Saiteja Utpala and
                  Zachary Giboney and
                  Gashaw M. Goshu and
                  Joan of Arc Xavier and
                  Sarah{-}Jane Crowson and
                  Mohinder Maheshbhai Naiya and
                  Noah Burns and
                  Lennart Finke and
                  Zerui Cheng and
                  Hyunwoo Park and
                  Francesco Fournier{-}Facio and
                  John Wydallis and
                  Mark Nandor and
                  Ankit Singh and
                  Tim Gehrunger and
                  Jiaqi Cai and
                  Ben McCarty and
                  Darling Duclosel and
                  Jungbae Nam and
                  Jennifer Zampese and
                  Ryan G. Hoerr and
                  Aras Bacho and
                  Gautier Abou Loume and
                  Abdallah Galal and
                  Hangrui Cao and
                  Alexis C. Garretson and
                  Damien Sileo and
                  Qiuyu Ren and
                  Doru Cojoc and
                  Pavel Arkhipov and
                  Usman Qazi and
                  Lianghui Li and
                  Sumeet Motwani and
                  Christian Schr{\"{o}}der de Witt and
                  Edwin Taylor and
                  Johannes Veith and
                  Eric Singer and
                  Taylor D. Hartman and
                  Paolo Rissone and
                  Jaehyeok Jin and
                  Jack Wei Lun Shi and
                  Chris G. Willcocks and
                  Joshua Robinson and
                  Aleksandar Mikov and
                  Ameya Prabhu and
                  Longke Tang and
                  Xavier Alapont and
                  Justine Leon Uro and
                  Kevin Zhou and
                  Emily de Oliveira Santos and
                  Andrey Pupasov Maksimov and
                  Edward Vendrow and
                  Kengo Zenitani and
                  Julien Guillod and
                  Yuqi Li and
                  Joshua Vendrow and
                  Vladyslav Kuchkin and
                  Ng Ze{-}An},
  title        = {Humanity's Last Exam},
  journal      = {CoRR},
  volume       = {abs/2501.14249},
  year         = {2025},
  url          = {https://doi.org/10.48550/arXiv.2501.14249},
  doi          = {10.48550/ARXIV.2501.14249},
  eprinttype    = {arXiv},
  eprint       = {2501.14249},
  bibsource    = {dblp computer science bibliography, https://dblp.org}
}

@inproceedings{Wei_CoT,
  author       = {Jason Wei and
                  Xuezhi Wang and
                  Dale Schuurmans and
                  Maarten Bosma and
                  Brian Ichter and
                  Fei Xia and
                  Ed H. Chi and
                  Quoc V. Le and
                  Denny Zhou},
  editor       = {Sanmi Koyejo and
                  S. Mohamed and
                  A. Agarwal and
                  Danielle Belgrave and
                  K. Cho and
                  A. Oh},
  title        = {Chain-of-Thought Prompting Elicits Reasoning in Large Language Models},
  booktitle    = {Advances in Neural Information Processing Systems 35: Annual Conference
                  on Neural Information Processing Systems 2022, NeurIPS 2022, New Orleans,
                  LA, USA, November 28 - December 9, 2022},
  year         = {2022},
  url          = {http://papers.nips.cc/paper\_files/paper/2022/hash/9d5609613524ecf4f15af0f7b31abca4-Abstract-Conference.html},
  bibsource    = {dblp computer science bibliography, https://dblp.org}
}

@misc{ToRL,
      title={ToRL: Scaling Tool-Integrated RL}, 
      author={Xuefeng Li and Haoyang Zou and Pengfei Liu},
      year={2025},
      eprint={2503.23383},
      archivePrefix={arXiv},
      primaryClass={cs.CL},
      url={https://arxiv.org/abs/2503.23383}, 
}

@misc{ReTool,
      title={ReTool: Reinforcement Learning for Strategic Tool Use in LLMs}, 
      author={Jiazhan Feng and Shijue Huang and Xingwei Qu and Ge Zhang and Yujia Qin and Baoquan Zhong and Chengquan Jiang and Jinxin Chi and Wanjun Zhong},
      year={2025},
      eprint={2504.11536},
      archivePrefix={arXiv},
      primaryClass={cs.CL},
      url={https://arxiv.org/abs/2504.11536}, 
}

@article{llm_agent_se,
  author       = {Haolin Jin and
                  Linghan Huang and
                  Haipeng Cai and
                  Jun Yan and
                  Bo Li and
                  Huaming Chen},
  title        = {From LLMs to LLM-based Agents for Software Engineering: {A} Survey
                  of Current, Challenges and Future},
  journal      = {CoRR},
  volume       = {abs/2408.02479},
  year         = {2024},
  url          = {https://doi.org/10.48550/arXiv.2408.02479},
  doi          = {10.48550/ARXIV.2408.02479},
  eprinttype    = {arXiv},
  eprint       = {2408.02479},
  bibsource    = {dblp computer science bibliography, https://dblp.org}
}

@article{tool_learning_survey,
  author       = {Changle Qu and
                  Sunhao Dai and
                  Xiaochi Wei and
                  Hengyi Cai and
                  Shuaiqiang Wang and
                  Dawei Yin and
                  Jun Xu and
                  Ji{-}Rong Wen},
  title        = {Tool learning with large language models: a survey},
  journal      = {Frontiers Comput. Sci.},
  volume       = {19},
  number       = {8},
  pages        = {198343},
  year         = {2025},
  url          = {https://doi.org/10.1007/s11704-024-40678-2},
  doi          = {10.1007/S11704-024-40678-2},
  bibsource    = {dblp computer science bibliography, https://dblp.org}
}

@inproceedings{Plan-and-Solve,
  author       = {Lei Wang and
                  Wanyu Xu and
                  Yihuai Lan and
                  Zhiqiang Hu and
                  Yunshi Lan and
                  Roy Ka{-}Wei Lee and
                  Ee{-}Peng Lim},
  editor       = {Anna Rogers and
                  Jordan L. Boyd{-}Graber and
                  Naoaki Okazaki},
  title        = {Plan-and-Solve Prompting: Improving Zero-Shot Chain-of-Thought Reasoning
                  by Large Language Models},
  booktitle    = {Proceedings of the 61st Annual Meeting of the Association for Computational
                  Linguistics (Volume 1: Long Papers), {ACL} 2023, Toronto, Canada,
                  July 9-14, 2023},
  pages        = {2609--2634},
  publisher    = {Association for Computational Linguistics},
  year         = {2023},
  url          = {https://doi.org/10.18653/v1/2023.acl-long.147},
  doi          = {10.18653/V1/2023.ACL-LONG.147},
  bibsource    = {dblp computer science bibliography, https://dblp.org}
}

@article{ARPO,
  author       = {Guanting Dong and
                  Hangyu Mao and
                  Kai Ma and
                  Licheng Bao and
                  Yifei Chen and
                  Zhongyuan Wang and
                  Zhongxia Chen and
                  Jiazhen Du and
                  Huiyang Wang and
                  Fuzheng Zhang and
                  Guorui Zhou and
                  Yutao Zhu and
                  Ji{-}Rong Wen and
                  Zhicheng Dou},
  title        = {Agentic Reinforced Policy Optimization},
  journal      = {CoRR},
  volume       = {abs/2507.19849},
  year         = {2025},
  url          = {https://doi.org/10.48550/arXiv.2507.19849},
  doi          = {10.48550/ARXIV.2507.19849},
  eprinttype    = {arXiv},
  eprint       = {2507.19849},
  bibsource    = {dblp computer science bibliography, https://dblp.org}
}

@inproceedings{DRAFT,
  author       = {Changle Qu and
                  Sunhao Dai and
                  Xiaochi Wei and
                  Hengyi Cai and
                  Shuaiqiang Wang and
                  Dawei Yin and
                  Jun Xu and
                  Ji{-}Rong Wen},
  title        = {From Exploration to Mastery: Enabling LLMs to Master Tools via Self-Driven
                  Interactions},
  booktitle    = {The Thirteenth International Conference on Learning Representations,
                  {ICLR} 2025, Singapore, April 24-28, 2025},
  publisher    = {OpenReview.net},
  year         = {2025},
  url          = {https://openreview.net/forum?id=QKBu1BOAwd},
  bibsource    = {dblp computer science bibliography, https://dblp.org}
}

@inproceedings{CodeAct,
  author       = {Xingyao Wang and
                  Yangyi Chen and
                  Lifan Yuan and
                  Yizhe Zhang and
                  Yunzhu Li and
                  Hao Peng and
                  Heng Ji},
  title        = {Executable Code Actions Elicit Better {LLM} Agents},
  booktitle    = {Forty-first International Conference on Machine Learning, {ICML} 2024,
                  Vienna, Austria, July 21-27, 2024},
  publisher    = {OpenReview.net},
  year         = {2024},
  url          = {https://openreview.net/forum?id=jJ9BoXAfFa},
  bibsource    = {dblp computer science bibliography, https://dblp.org}
}

@article{HiRA,
  author       = {Jiajie Jin and
                  Xiaoxi Li and
                  Guanting Dong and
                  Yuyao Zhang and
                  Yutao Zhu and
                  Zhao Yang and
                  Hongjin Qian and
                  Zhicheng Dou},
  title        = {Decoupled Planning and Execution: {A} Hierarchical Reasoning Framework
                  for Deep Search},
  journal      = {CoRR},
  volume       = {abs/2507.02652},
  year         = {2025},
  url          = {https://doi.org/10.48550/arXiv.2507.02652},
  doi          = {10.48550/ARXIV.2507.02652},
  eprinttype    = {arXiv},
  eprint       = {2507.02652},
  bibsource    = {dblp computer science bibliography, https://dblp.org}
}

@misc{AgentScaler,
      title={Towards General Agentic Intelligence via Environment Scaling}, 
      author={Runnan Fang and Shihao Cai and Baixuan Li and Jialong Wu and Guangyu Li and Wenbiao Yin and Xinyu Wang and Xiaobin Wang and Liangcai Su and Zhen Zhang and Shibin Wu and Zhengwei Tao and Yong Jiang and Pengjun Xie and Fei Huang and Jingren Zhou},
      year={2025},
      eprint={2509.13311},
      archivePrefix={arXiv},
      primaryClass={cs.CL},
      url={https://arxiv.org/abs/2509.13311}, 
}

@misc{VerlTool,
      title={VerlTool: Towards Holistic Agentic Reinforcement Learning with Tool Use}, 
      author={Dongfu Jiang and Yi Lu and Zhuofeng Li and Zhiheng Lyu and Ping Nie and Haozhe Wang and Alex Su and Hui Chen and Kai Zou and Chao Du and Tianyu Pang and Wenhu Chen},
      year={2025},
      eprint={2509.01055},
      archivePrefix={arXiv},
      primaryClass={cs.AI},
      url={https://arxiv.org/abs/2509.01055}, 
}

@inproceedings{ToolLLM,
  author       = {Yujia Qin and
                  Shihao Liang and
                  Yining Ye and
                  Kunlun Zhu and
                  Lan Yan and
                  Yaxi Lu and
                  Yankai Lin and
                  Xin Cong and
                  Xiangru Tang and
                  Bill Qian and
                  Sihan Zhao and
                  Lauren Hong and
                  Runchu Tian and
                  Ruobing Xie and
                  Jie Zhou and
                  Mark Gerstein and
                  Dahai Li and
                  Zhiyuan Liu and
                  Maosong Sun},
  title        = {ToolLLM: Facilitating Large Language Models to Master 16000+ Real-world
                  APIs},
  booktitle    = {The Twelfth International Conference on Learning Representations,
                  {ICLR} 2024, Vienna, Austria, May 7-11, 2024},
  publisher    = {OpenReview.net},
  year         = {2024},
  url          = {https://openreview.net/forum?id=dHng2O0Jjr},
  bibsource    = {dblp computer science bibliography, https://dblp.org}
}

@article{MCP,
  author       = {Xinyi Hou and
                  Yanjie Zhao and
                  Shenao Wang and
                  Haoyu Wang},
  title        = {Model Context Protocol {(MCP):} Landscape, Security Threats, and Future
                  Research Directions},
  journal      = {CoRR},
  volume       = {abs/2503.23278},
  year         = {2025},
  url          = {https://doi.org/10.48550/arXiv.2503.23278},
  doi          = {10.48550/ARXIV.2503.23278},
  eprinttype    = {arXiv},
  eprint       = {2503.23278},
  bibsource    = {dblp computer science bibliography, https://dblp.org}
}

@inproceedings{ToolRet,
  author       = {Zhengliang Shi and
                  Yuhan Wang and
                  Lingyong Yan and
                  Pengjie Ren and
                  Shuaiqiang Wang and
                  Dawei Yin and
                  Zhaochun Ren},
  editor       = {Wanxiang Che and
                  Joyce Nabende and
                  Ekaterina Shutova and
                  Mohammad Taher Pilehvar},
  title        = {Retrieval Models Aren't Tool-Savvy: Benchmarking Tool Retrieval for
                  Large Language Models},
  booktitle    = {Findings of the Association for Computational Linguistics, {ACL} 2025,
                  Vienna, Austria, July 27 - August 1, 2025},
  pages        = {24497--24524},
  publisher    = {Association for Computational Linguistics},
  year         = {2025},
  url          = {https://aclanthology.org/2025.findings-acl.1258/},
  bibsource    = {dblp computer science bibliography, https://dblp.org}
}

@misc{LIMI,
      title={LIMI: Less is More for Agency}, 
      author={Yang Xiao and Mohan Jiang and Jie Sun and Keyu Li and Jifan Lin and Yumin Zhuang and Ji Zeng and Shijie Xia and Qishuo Hua and Xuefeng Li and Xiaojie Cai and Tongyu Wang and Yue Zhang and Liming Liu and Xia Wu and Jinlong Hou and Yuan Cheng and Wenjie Li and Xiang Wang and Dequan Wang and Pengfei Liu},
      year={2025},
      eprint={2509.17567},
      archivePrefix={arXiv},
      primaryClass={cs.AI},
      url={https://arxiv.org/abs/2509.17567}, 
}

@inproceedings{ToolHop,
  author       = {Junjie Ye and
                  Zhengyin Du and
                  Xuesong Yao and
                  Weijian Lin and
                  Yufei Xu and
                  Zehui Chen and
                  Zaiyuan Wang and
                  Sining Zhu and
                  Zhiheng Xi and
                  Siyu Yuan and
                  Tao Gui and
                  Qi Zhang and
                  Xuanjing Huang and
                  Jiecao Chen},
  editor       = {Wanxiang Che and
                  Joyce Nabende and
                  Ekaterina Shutova and
                  Mohammad Taher Pilehvar},
  title        = {ToolHop: {A} Query-Driven Benchmark for Evaluating Large Language
                  Models in Multi-Hop Tool Use},
  booktitle    = {Proceedings of the 63rd Annual Meeting of the Association for Computational
                  Linguistics (Volume 1: Long Papers), {ACL} 2025, Vienna, Austria,
                  July 27 - August 1, 2025},
  pages        = {2995--3021},
  publisher    = {Association for Computational Linguistics},
  year         = {2025},
  url          = {https://aclanthology.org/2025.acl-long.150/},
  bibsource    = {dblp computer science bibliography, https://dblp.org}
}

@inproceedings{WebShop,
  author       = {Shunyu Yao and
                  Howard Chen and
                  John Yang and
                  Karthik Narasimhan},
  editor       = {Sanmi Koyejo and
                  S. Mohamed and
                  A. Agarwal and
                  Danielle Belgrave and
                  K. Cho and
                  A. Oh},
  title        = {WebShop: Towards Scalable Real-World Web Interaction with Grounded
                  Language Agents},
  booktitle    = {Advances in Neural Information Processing Systems 35: Annual Conference
                  on Neural Information Processing Systems 2022, NeurIPS 2022, New Orleans,
                  LA, USA, November 28 - December 9, 2022},
  year         = {2022},
  url          = {http://papers.nips.cc/paper\_files/paper/2022/hash/82ad13ec01f9fe44c01cb91814fd7b8c-Abstract-Conference.html},
  bibsource    = {dblp computer science bibliography, https://dblp.org}
}

@inproceedings{ALFWorld,
  author       = {Mohit Shridhar and
                  Xingdi Yuan and
                  Marc{-}Alexandre C{\^{o}}t{\'{e}} and
                  Yonatan Bisk and
                  Adam Trischler and
                  Matthew J. Hausknecht},
  title        = {ALFWorld: Aligning Text and Embodied Environments for Interactive
                  Learning},
  booktitle    = {9th International Conference on Learning Representations, {ICLR} 2021,
                  Virtual Event, Austria, May 3-7, 2021},
  publisher    = {OpenReview.net},
  year         = {2021},
  url          = {https://openreview.net/forum?id=0IOX0YcCdTn},
  bibsource    = {dblp computer science bibliography, https://dblp.org}
}

@article{RestGPT,
  author       = {Yifan Song and
                  Weimin Xiong and
                  Dawei Zhu and
                  Cheng Li and
                  Ke Wang and
                  Ye Tian and
                  Sujian Li},
  title        = {RestGPT: Connecting Large Language Models with Real-World Applications
                  via RESTful APIs},
  journal      = {CoRR},
  volume       = {abs/2306.06624},
  year         = {2023},
  url          = {https://doi.org/10.48550/arXiv.2306.06624},
  doi          = {10.48550/ARXIV.2306.06624},
  eprinttype    = {arXiv},
  eprint       = {2306.06624},
  bibsource    = {dblp computer science bibliography, https://dblp.org}
}

@inproceedings{API-Bank,
  author       = {Minghao Li and
                  Yingxiu Zhao and
                  Bowen Yu and
                  Feifan Song and
                  Hangyu Li and
                  Haiyang Yu and
                  Zhoujun Li and
                  Fei Huang and
                  Yongbin Li},
  editor       = {Houda Bouamor and
                  Juan Pino and
                  Kalika Bali},
  title        = {API-Bank: {A} Comprehensive Benchmark for Tool-Augmented LLMs},
  booktitle    = {Proceedings of the 2023 Conference on Empirical Methods in Natural
                  Language Processing, {EMNLP} 2023, Singapore, December 6-10, 2023},
  pages        = {3102--3116},
  publisher    = {Association for Computational Linguistics},
  year         = {2023},
  url          = {https://doi.org/10.18653/v1/2023.emnlp-main.187},
  doi          = {10.18653/V1/2023.EMNLP-MAIN.187},
  bibsource    = {dblp computer science bibliography, https://dblp.org}
}

@article{WebDancer,
  author       = {Jialong Wu and
                  Baixuan Li and
                  Runnan Fang and
                  Wenbiao Yin and
                  Liwen Zhang and
                  Zhengwei Tao and
                  Dingchu Zhang and
                  Zekun Xi and
                  Yong Jiang and
                  Pengjun Xie and
                  Fei Huang and
                  Jingren Zhou},
  title        = {WebDancer: Towards Autonomous Information Seeking Agency},
  journal      = {CoRR},
  volume       = {abs/2505.22648},
  year         = {2025},
  url          = {https://doi.org/10.48550/arXiv.2505.22648},
  doi          = {10.48550/ARXIV.2505.22648},
  eprinttype    = {arXiv},
  eprint       = {2505.22648},
  bibsource    = {dblp computer science bibliography, https://dblp.org}
}

@article{WebShaper,
  author       = {Zhengwei Tao and
                  Jialong Wu and
                  Wenbiao Yin and
                  Junkai Zhang and
                  Baixuan Li and
                  Haiyang Shen and
                  Kuan Li and
                  Liwen Zhang and
                  Xinyu Wang and
                  Yong Jiang and
                  Pengjun Xie and
                  Fei Huang and
                  Jingren Zhou},
  title        = {WebShaper: Agentically Data Synthesizing via Information-Seeking Formalization},
  journal      = {CoRR},
  volume       = {abs/2507.15061},
  year         = {2025},
  url          = {https://doi.org/10.48550/arXiv.2507.15061},
  doi          = {10.48550/ARXIV.2507.15061},
  eprinttype    = {arXiv},
  eprint       = {2507.15061},
  bibsource    = {dblp computer science bibliography, https://dblp.org}
}

@article{DeepMath,
  title={DeepMath-103K: A Large-Scale, Challenging, Decontaminated, and  Verifiable Mathematical Dataset for Advancing Reasoning},
  author={He, Zhiwei and Liang, Tian and Xu, Jiahao and Liu, Qiuzhi and Chen, Xingyu and Wang, Yue and Song, Linfeng and Yu, Dian and Liang, Zhenwen and Wang, Wenxuan and Zhang, Zhuosheng and Wang, Rui and Tu, Zhaopeng and Mi, Haitao and Yu, Dong},
  year={2025},
  eprint={2504.11456},
  archivePrefix={arXiv},
  primaryClass={cs.CL},
  url={https://arxiv.org/abs/2504.11456}, 
}

@article{WebThinker,
  author       = {Xiaoxi Li and
                  Jiajie Jin and
                  Guanting Dong and
                  Hongjin Qian and
                  Yutao Zhu and
                  Yongkang Wu and
                  Ji{-}Rong Wen and
                  Zhicheng Dou},
  title        = {WebThinker: Empowering Large Reasoning Models with Deep Research Capability},
  journal      = {CoRR},
  volume       = {abs/2504.21776},
  year         = {2025},
  url          = {https://doi.org/10.48550/arXiv.2504.21776},
  doi          = {10.48550/ARXIV.2504.21776},
  eprinttype    = {arXiv},
  eprint       = {2504.21776},
  bibsource    = {dblp computer science bibliography, https://dblp.org}
}

@article{Qwen3,
  author       = {An Yang and
                  Anfeng Li and
                  Baosong Yang and
                  Beichen Zhang and
                  Binyuan Hui and
                  Bo Zheng and
                  Bowen Yu and
                  Chang Gao and
                  Chengen Huang and
                  Chenxu Lv and
                  Chujie Zheng and
                  Dayiheng Liu and
                  Fan Zhou and
                  Fei Huang and
                  Feng Hu and
                  Hao Ge and
                  Haoran Wei and
                  Huan Lin and
                  Jialong Tang and
                  Jian Yang and
                  Jianhong Tu and
                  Jianwei Zhang and
                  Jian Yang and
                  Jiaxi Yang and
                  Jingren Zhou and
                  Junyang Lin and
                  Kai Dang and
                  Keqin Bao and
                  Kexin Yang and
                  Le Yu and
                  Lianghao Deng and
                  Mei Li and
                  Mingfeng Xue and
                  Mingze Li and
                  Pei Zhang and
                  Peng Wang and
                  Qin Zhu and
                  Rui Men and
                  Ruize Gao and
                  Shixuan Liu and
                  Shuang Luo and
                  Tianhao Li and
                  Tianyi Tang and
                  Wenbiao Yin and
                  Xingzhang Ren and
                  Xinyu Wang and
                  Xinyu Zhang and
                  Xuancheng Ren and
                  Yang Fan and
                  Yang Su and
                  Yichang Zhang and
                  Yinger Zhang and
                  Yu Wan and
                  Yuqiong Liu and
                  Zekun Wang and
                  Zeyu Cui and
                  Zhenru Zhang and
                  Zhipeng Zhou and
                  Zihan Qiu},
  title        = {Qwen3 Technical Report},
  journal      = {CoRR},
  volume       = {abs/2505.09388},
  year         = {2025},
  url          = {https://doi.org/10.48550/arXiv.2505.09388},
  doi          = {10.48550/ARXIV.2505.09388},
  eprinttype    = {arXiv},
  eprint       = {2505.09388},
  bibsource    = {dblp computer science bibliography, https://dblp.org}
}

@inproceedings{C-Pack,
  author       = {Shitao Xiao and
                  Zheng Liu and
                  Peitian Zhang and
                  Niklas Muennighoff and
                  Defu Lian and
                  Jian{-}Yun Nie},
  editor       = {Grace Hui Yang and
                  Hongning Wang and
                  Sam Han and
                  Claudia Hauff and
                  Guido Zuccon and
                  Yi Zhang},
  title        = {C-Pack: Packed Resources For General Chinese Embeddings},
  booktitle    = {Proceedings of the 47th International {ACM} {SIGIR} Conference on
                  Research and Development in Information Retrieval, {SIGIR} 2024, Washington
                  DC, USA, July 14-18, 2024},
  pages        = {641--649},
  publisher    = {{ACM}},
  year         = {2024},
  url          = {https://doi.org/10.1145/3626772.3657878},
  doi          = {10.1145/3626772.3657878},
  bibsource    = {dblp computer science bibliography, https://dblp.org}
}

@inproceedings{AgentTuning,
  author       = {Aohan Zeng and
                  Mingdao Liu and
                  Rui Lu and
                  Bowen Wang and
                  Xiao Liu and
                  Yuxiao Dong and
                  Jie Tang},
  editor       = {Lun{-}Wei Ku and
                  Andre Martins and
                  Vivek Srikumar},
  title        = {AgentTuning: Enabling Generalized Agent Abilities for LLMs},
  booktitle    = {Findings of the Association for Computational Linguistics, {ACL} 2024,
                  Bangkok, Thailand and virtual meeting, August 11-16, 2024},
  pages        = {3053--3077},
  publisher    = {Association for Computational Linguistics},
  year         = {2024},
  url          = {https://doi.org/10.18653/v1/2024.findings-acl.181},
  doi          = {10.18653/V1/2024.FINDINGS-ACL.181},
  bibsource    = {dblp computer science bibliography, https://dblp.org}
}

@inproceedings{Reflexion,
  author       = {Noah Shinn and
                  Federico Cassano and
                  Ashwin Gopinath and
                  Karthik Narasimhan and
                  Shunyu Yao},
  editor       = {Alice Oh and
                  Tristan Naumann and
                  Amir Globerson and
                  Kate Saenko and
                  Moritz Hardt and
                  Sergey Levine},
  title        = {Reflexion: language agents with verbal reinforcement learning},
  booktitle    = {Advances in Neural Information Processing Systems 36: Annual Conference
                  on Neural Information Processing Systems 2023, NeurIPS 2023, New Orleans,
                  LA, USA, December 10 - 16, 2023},
  year         = {2023},
  url          = {http://papers.nips.cc/paper\_files/paper/2023/hash/1b44b878bb782e6954cd888628510e90-Abstract-Conference.html},
  bibsource    = {dblp computer science bibliography, https://dblp.org}
}

@article{verl,
  title   = {HybridFlow: A Flexible and Efficient RLHF Framework},
  author  = {Guangming Sheng and Chi Zhang and Zilingfeng Ye and Xibin Wu and Wang Zhang and Ru Zhang and Yanghua Peng and Haibin Lin and Chuan Wu},
  year    = {2024},
  journal = {arXiv preprint arXiv: 2409.19256}
}

@misc{ACON,
      title={ACON: Optimizing Context Compression for Long-horizon LLM Agents}, 
      author={Minki Kang and Wei-Ning Chen and Dongge Han and Huseyin A. Inan and Lukas Wutschitz and Yanzhi Chen and Robert Sim and Saravan Rajmohan},
      year={2025},
      eprint={2510.00615},
      archivePrefix={arXiv},
      primaryClass={cs.AI},
      url={https://arxiv.org/abs/2510.00615}, 
}

@misc{FoldGRPO,
      title={Scaling Long-Horizon LLM Agent via Context-Folding}, 
      author={Weiwei Sun and Miao Lu and Zhan Ling and Kang Liu and Xuesong Yao and Yiming Yang and Jiecao Chen},
      year={2025},
      eprint={2510.11967},
      archivePrefix={arXiv},
      primaryClass={cs.CL},
      url={https://arxiv.org/abs/2510.11967}, 
}

@misc{AEPO,
      title={Agentic Entropy-Balanced Policy Optimization}, 
      author={Guanting Dong and Licheng Bao and Zhongyuan Wang and Kangzhi Zhao and Xiaoxi Li and Jiajie Jin and Jinghan Yang and Hangyu Mao and Fuzheng Zhang and Kun Gai and Guorui Zhou and Yutao Zhu and Ji-Rong Wen and Zhicheng Dou},
      year={2025},
      eprint={2510.14545},
      archivePrefix={arXiv},
      primaryClass={cs.LG},
      url={https://arxiv.org/abs/2510.14545}, 
}

@misc{ToolLight,
      title={Toward Effective Tool-Integrated Reasoning via Self-Evolved Preference Learning}, 
      author={Yifei Chen and Guanting Dong and Zhicheng Dou},
      year={2025},
      eprint={2509.23285},
      archivePrefix={arXiv},
      primaryClass={cs.AI},
      url={https://arxiv.org/abs/2509.23285}, 
}

@inproceedings{ToolGen,
  author       = {Renxi Wang and
                  Xudong Han and
                  Lei Ji and
                  Shu Wang and
                  Timothy Baldwin and
                  Haonan Li},
  title        = {ToolGen: Unified Tool Retrieval and Calling via Generation},
  booktitle    = {The Thirteenth International Conference on Learning Representations,
                  {ICLR} 2025, Singapore, April 24-28, 2025},
  publisher    = {OpenReview.net},
  year         = {2025},
  url          = {https://openreview.net/forum?id=XLMAMmowdY},
  bibsource    = {dblp computer science bibliography, https://dblp.org}
}

@article{Tool-Star,
  author       = {Guanting Dong and
                  Yifei Chen and
                  Xiaoxi Li and
                  Jiajie Jin and
                  Hongjin Qian and
                  Yutao Zhu and
                  Hangyu Mao and
                  Guorui Zhou and
                  Zhicheng Dou and
                  Ji{-}Rong Wen},
  title        = {Tool-Star: Empowering LLM-Brained Multi-Tool Reasoner via Reinforcement
                  Learning},
  journal      = {CoRR},
  volume       = {abs/2505.16410},
  year         = {2025},
  url          = {https://doi.org/10.48550/arXiv.2505.16410},
  doi          = {10.48550/ARXIV.2505.16410},
  eprinttype    = {arXiv},
  eprint       = {2505.16410},
  bibsource    = {dblp computer science bibliography, https://dblp.org}
}

@misc{FinSight,
      title={FinSight: Towards Real-World Financial Deep Research}, 
      author={Jiajie Jin and Yuyao Zhang and Yimeng Xu and Hongjin Qian and Yutao Zhu and Zhicheng Dou},
      year={2025},
      eprint={2510.16844},
      archivePrefix={arXiv},
      primaryClass={cs.CL},
      url={https://arxiv.org/abs/2510.16844}, 
}

@article{RAGEN,
  author       = {Zihan Wang and
                  Kangrui Wang and
                  Qineng Wang and
                  Pingyue Zhang and
                  Linjie Li and
                  Zhengyuan Yang and
                  Xing Jin and
                  Kefan Yu and
                  Minh Nhat Nguyen and
                  Licheng Liu and
                  Eli Gottlieb and
                  Yiping Lu and
                  Kyunghyun Cho and
                  Jiajun Wu and
                  Li Fei{-}Fei and
                  Lijuan Wang and
                  Yejin Choi and
                  Manling Li},
  title        = {{RAGEN:} Understanding Self-Evolution in {LLM} Agents via Multi-Turn
                  Reinforcement Learning},
  journal      = {CoRR},
  volume       = {abs/2504.20073},
  year         = {2025},
  url          = {https://doi.org/10.48550/arXiv.2504.20073},
  doi          = {10.48550/ARXIV.2504.20073},
  eprinttype    = {arXiv},
  eprint       = {2504.20073},
  bibsource    = {dblp computer science bibliography, https://dblp.org}
}

@article{WebExplorer,
  author       = {Junteng Liu and
                  Yunji Li and
                  Chi Zhang and
                  Jingyang Li and
                  Aili Chen and
                  Ke Ji and
                  Weiyu Cheng and
                  Zijia Wu and
                  Chengyu Du and
                  Qidi Xu and
                  Jiayuan Song and
                  Zhengmao Zhu and
                  Wenhu Chen and
                  Pengyu Zhao and
                  Junxian He},
  title        = {WebExplorer: Explore and Evolve for Training Long-Horizon Web Agents},
  journal      = {CoRR},
  volume       = {abs/2509.06501},
  year         = {2025},
  url          = {https://doi.org/10.48550/arXiv.2509.06501},
  doi          = {10.48550/ARXIV.2509.06501},
  eprinttype    = {arXiv},
  eprint       = {2509.06501},
  bibsource    = {dblp computer science bibliography, https://dblp.org}
}

@article{Thinking-Optimal,
  author       = {Wenkai Yang and
                  Shuming Ma and
                  Yankai Lin and
                  Furu Wei},
  title        = {Towards Thinking-Optimal Scaling of Test-Time Compute for {LLM} Reasoning},
  journal      = {CoRR},
  volume       = {abs/2502.18080},
  year         = {2025},
  url          = {https://doi.org/10.48550/arXiv.2502.18080},
  doi          = {10.48550/ARXIV.2502.18080},
  eprinttype    = {arXiv},
  eprint       = {2502.18080},
  bibsource    = {dblp computer science bibliography, https://dblp.org}
}

@misc{GEM,
      title={GEM: A Gym for Agentic LLMs}, 
      author={Zichen Liu and Anya Sims and Keyu Duan and Changyu Chen and Simon Yu and Xiangxin Zhou and Haotian Xu and Shaopan Xiong and Bo Liu and Chenmien Tan and Chuen Yang Beh and Weixun Wang and Hao Zhu and Weiyan Shi and Diyi Yang and Michael Shieh and Yee Whye Teh and Wee Sun Lee and Min Lin},
      year={2025},
      eprint={2510.01051},
      archivePrefix={arXiv},
      primaryClass={cs.LG},
      url={https://arxiv.org/abs/2510.01051}, 
}

@misc{TourPlanner,
      title={TourPlanner: A Competitive Consensus Framework with Constraint-Gated Reinforcement Learning for Travel Planning}, 
      author={Yinuo Wang and Mining Tan and Wenxiang Jiao and Xiaoxi Li and Hao Wang and Xuanyu Zhang and Yuan Lu and Weiming Dong},
      year={2026},
      eprint={2601.04698},
      archivePrefix={arXiv},
      primaryClass={cs.AI},
      url={https://arxiv.org/abs/2601.04698}, 
}

@misc{Diffusion-RL,
      title={Diffusion Actor-Critic with Entropy Regulator}, 
      author={Yinuo Wang and Likun Wang and Yuxuan Jiang and Wenjun Zou and Tong Liu and Xujie Song and Wenxuan Wang and Liming Xiao and Jiang Wu and Jingliang Duan and Shengbo Eben Li},
      year={2024},
      eprint={2405.15177},
      archivePrefix={arXiv},
      primaryClass={cs.LG},
      url={https://arxiv.org/abs/2405.15177}, 
}

@misc{CompactClueSelection-RAG,
      title={Less is More: Compact Clue Selection for Efficient Retrieval-Augmented Generation Reasoning}, 
      author={Qianchi Zhang and Hainan Zhang and Liang Pang and Hongwei Zheng and Yongxin Tong and Zhiming Zheng},
      year={2025},
      eprint={2502.11811},
      archivePrefix={arXiv},
      primaryClass={cs.CL},
      url={https://arxiv.org/abs/2502.11811}, 
}

@misc{AgentGen,
      title={AgentGen: Enhancing Planning Abilities for Large Language Model based Agent via Environment and Task Generation}, 
      author={Mengkang Hu and Pu Zhao and Can Xu and Qingfeng Sun and Jianguang Lou and Qingwei Lin and Ping Luo and Saravan Rajmohan},
      year={2025},
      eprint={2408.00764},
      archivePrefix={arXiv},
      primaryClass={cs.CL},
      url={https://arxiv.org/abs/2408.00764}, 
}

@misc{HiAgent,
      title={HiAgent: Hierarchical Working Memory Management for Solving Long-Horizon Agent Tasks with Large Language Model}, 
      author={Mengkang Hu and Tianxing Chen and Qiguang Chen and Yao Mu and Wenqi Shao and Ping Luo},
      year={2024},
      eprint={2408.09559},
      archivePrefix={arXiv},
      primaryClass={cs.CL},
      url={https://arxiv.org/abs/2408.09559}, 
}

@misc{Agent2World,
      title={Agent2World: Learning to Generate Symbolic World Models via Adaptive Multi-Agent Feedback}, 
      author={Mengkang Hu and Bowei Xia and Yuran Wu and Ailing Yu and Yude Zou and Qiguang Chen and Shijian Wang and Jiarui Jin and Kexin Li and Wenxiang Jiao and Yuan Lu and Ping Luo},
      year={2025},
      eprint={2512.22336},
      archivePrefix={arXiv},
      primaryClass={cs.AI},
      url={https://arxiv.org/abs/2512.22336}, 
}

@misc{EnvScaler,
      title={EnvScaler: Scaling Tool-Interactive Environments for LLM Agent via Programmatic Synthesis}, 
      author={Xiaoshuai Song and Haofei Chang and Guanting Dong and Yutao Zhu and Zhicheng Dou and Ji-Rong Wen},
      year={2026},
      eprint={2601.05808},
      archivePrefix={arXiv},
      primaryClass={cs.CL},
      url={https://arxiv.org/abs/2601.05808}, 
}

@misc{HierSearch,
      title={HierSearch: A Hierarchical Enterprise Deep Search Framework Integrating Local and Web Searches}, 
      author={Jiejun Tan and Zhicheng Dou and Yan Yu and Jiehan Cheng and Qiang Ju and Jian Xie and Ji-Rong Wen},
      year={2025},
      eprint={2508.08088},
      archivePrefix={arXiv},
      primaryClass={cs.IR},
      url={https://arxiv.org/abs/2508.08088}, 
}

@inproceedings{RetroLLM,
  author       = {Xiaoxi Li and
                  Jiajie Jin and
                  Yujia Zhou and
                  Yongkang Wu and
                  Zhonghua Li and
                  Ye Qi and
                  Zhicheng Dou},
  editor       = {Wanxiang Che and
                  Joyce Nabende and
                  Ekaterina Shutova and
                  Mohammad Taher Pilehvar},
  title        = {RetroLLM: Empowering Large Language Models to Retrieve Fine-grained
                  Evidence within Generation},
  booktitle    = {Proceedings of the 63rd Annual Meeting of the Association for Computational
                  Linguistics (Volume 1: Long Papers), {ACL} 2025, Vienna, Austria,
                  July 27 - August 1, 2025},
  pages        = {16754--16779},
  publisher    = {Association for Computational Linguistics},
  year         = {2025},
  url          = {https://aclanthology.org/2025.acl-long.819/},
  bibsource    = {dblp computer science bibliography, https://dblp.org}
}

@inproceedings{RAG-Critic,
  author       = {Guanting Dong and
                  Jiajie Jin and
                  Xiaoxi Li and
                  Yutao Zhu and
                  Zhicheng Dou and
                  Ji{-}Rong Wen},
  editor       = {Wanxiang Che and
                  Joyce Nabende and
                  Ekaterina Shutova and
                  Mohammad Taher Pilehvar},
  title        = {RAG-Critic: Leveraging Automated Critic-Guided Agentic Workflow for
                  Retrieval Augmented Generation},
  booktitle    = {Proceedings of the 63rd Annual Meeting of the Association for Computational
                  Linguistics (Volume 1: Long Papers), {ACL} 2025, Vienna, Austria,
                  July 27 - August 1, 2025},
  pages        = {3551--3578},
  publisher    = {Association for Computational Linguistics},
  year         = {2025},
  url          = {https://aclanthology.org/2025.acl-long.179/},
  bibsource    = {dblp computer science bibliography, https://dblp.org}
}

@inproceedings{LongRefiner,
  author       = {Jiajie Jin and
                  Xiaoxi Li and
                  Guanting Dong and
                  Yuyao Zhang and
                  Yutao Zhu and
                  Yongkang Wu and
                  Zhonghua Li and
                  Ye Qi and
                  Zhicheng Dou},
  editor       = {Wanxiang Che and
                  Joyce Nabende and
                  Ekaterina Shutova and
                  Mohammad Taher Pilehvar},
  title        = {Hierarchical Document Refinement for Long-context Retrieval-augmented
                  Generation},
  booktitle    = {Proceedings of the 63rd Annual Meeting of the Association for Computational
                  Linguistics (Volume 1: Long Papers), {ACL} 2025, Vienna, Austria,
                  July 27 - August 1, 2025},
  pages        = {3502--3520},
  publisher    = {Association for Computational Linguistics},
  year         = {2025},
  url          = {https://aclanthology.org/2025.acl-long.176/},
  bibsource    = {dblp computer science bibliography, https://dblp.org}
}

@article{Video-Thinker,
  title={Video-Thinker: Sparking" Thinking with Videos" via Reinforcement Learning},
  author={Wang, Shijian and Jin, Jiarui and Wang, Xingjian and Song, Linxin and Fu, Runhao and Wang, Hecheng and Ge, Zongyuan and Lu, Yuan and Cheng, Xuelian},
  journal={arXiv preprint arXiv:2510.23473},
  year={2025}
}

@article{SyntheticCurriculum,
  title={Synthetic Curriculum Reinforces Compositional Text-to-Image Generation},
  author={Wang, Shijian and Fu, Runhao and Zhao, Siyi and Zhan, Qingqin and Wang, Xingjian and Jin, Jiarui and Lu, Yuan and Wu, Hanqian and Chen, Cunjian},
  journal={arXiv preprint arXiv:2511.18378},
  year={2025}
}

% \newpage
\section*{Appendix}
\appendix

\section{Datasets}
\label{app:datasets}

\subsection{Training Data}
\label{app:training_data_collection}
We collected a diverse training dataset spanning four task categories to instill comprehensive agent capabilities.
\begin{itemize}[leftmargin=1em, nosep]
    \item \textbf{General Tool-Use}: We sample 1k instances for labeled-tool scenarios and 1k for tool-retrieval from the ToolBench~\cite{ToolLLM} training set. This data is intended to instill a generalized ability to use diverse tools and leverage large toolsets through retrieval.
    \item \textbf{Real-World Interaction}: We utilize 500 instances from ALFWorld~\cite{ALFWorld} and 500 from WebShop~\cite{WebShop}, sampled from their training sets, to teach the model to interact effectively with environments, manage state transitions, and achieve user goals.
    \item \textbf{Deep Research}: We include 200 instances from WebDancer~\cite{WebDancer} and 500 from WebShaperQA~\cite{WebShaper} to enhance the model's proficiency in using web search and page browsing for in-depth information gathering.
    \item \textbf{Mathematical Reasoning}: We collect 0.9k problems from the DeepMath dataset~\cite{DeepMath} to strengthen the model's ability to use code as a tool for complex mathematical computations.
\end{itemize}

\subsection{Benchmarks}
\label{app:benchmarks}
We conduct extensive experiments on a wide range of benchmarks, including general tool-use and downstream applications.
\paragraph{General Tool-Use}
These benchmarks encompass a broad range of distinct tools (from tens to over ten thousand), thus offering a testbed for evaluating different approaches to toolset scaling.
\begin{itemize}[leftmargin=1em, nosep]
    \item \textbf{ToolBench~\cite{ToolLLM}}: A large-scale benchmark containing over 16,000 real-world REST APIs spanning 49 categories. Test subsets include 100 test cases, designed to evaluate LLMs in both single-tool and complex multi-tool scenarios.
    \item \textbf{API-Bank~\cite{API-Bank}}: A comprehensive benchmark for tool-augmented LLMs. It features a runnable evaluation system with 73 API tools and a large training set (over 2,200 dialogues across 2,211 APIs from 1,008 domains), assessing LLMs' capabilities in planning, retrieving, and calling APIs.
    \item \textbf{TMDB~\cite{RestGPT}}: A sub-scenario of RestBench focused on the TMDB movie database, consisting of 100 questions that utilize 54 local tools and require an average of 2.3 sequential API calls.
    \item \textbf{Spotify~\cite{RestGPT}}: A sub-scenario of RestBench simulating a Spotify music player, featuring 57 questions and 40 local tools, demanding an average of 2.6 sequential API calls to complete the tasks.
    \item \textbf{ToolHop~\cite{ToolHop}}: A multi-hop reasoning dataset comprising 995 complex questions. It leverages 3,912 locally executable tools and requires between 3 to 7 sequential tool calls per task.
\end{itemize}
\paragraph{Downstream Applications} These benchmarks test the capability of different approaches in handling complex real-world tasks, which often require the use of domain-specific toolsets.
\begin{itemize}[leftmargin=1em, nosep]
    \item \textbf{ALFWorld~\cite{ALFWorld}}: A benchmark for simple Embodied AI tasks set in a text environment. Agents must complete objectives using a finite set of low-level embodied actions (eg., move, take) to test navigation and object manipulation.
    \item \textbf{WebShop~\cite{WebShop}}: A challenging online shopping environment that provides 12,087 crowd-sourced tasks over a catalog of 1.18 million products. Agents interact with the simulated e-commerce website using core APIs: search[Query] and choose[Text Button].
    \item \textbf{GAIA~\cite{GAIA}}: A complex benchmark for General AI Assistants, consisting of 466 real-world questions (with a 300-question held-out test set). It requires the flexible application of a broad general-purpose toolset including web browsing, code execution, multi-modal processing, and file handling.
    \item \textbf{Humanity's Last Exam~(HLE)~\cite{HLE}}: A benchmark featuring 2,500 highly difficult, multi-disciplinary questions (graduate-level). It primarily evaluates the model's intrinsic deep reasoning and multi-modal understanding capabilities, as the questions are designed to be insoluble by simple external search tools.
\end{itemize}

\section{Implementation Details}
\label{app:implementation_details}
For DeepAgent, we use QwQ-32B~\cite{qwen_qwq} as the main reasoning model for the results in Table~\ref{tab:tool_usage_performance} and Table~\ref{tab:application_preformance}, and Qwen3-30B-A3B-Thinking-2507~\cite{Qwen3} with Qwen3-235B-A22B-Thinking-2507~\cite{Qwen3} in Table~\ref{tab:backbone_comparison}. We use Qwen2.5-32B-Instruct~\cite{qwen2.5} as the auxiliary model for (1) filtering lengthy tool search results and execution outputs (this is also applied to all baselines), (2) simulating RapidAPIs during ToolPO training, and (3) generating folded memory from interaction history. For the baselines, we use either QwQ-32B or Qwen2.5-32B-Instruct as the backbone model. Text generation for all models uses a maximum of 81,920 tokens, with a temperature of 0.7, top\_p of 0.8, top\_k of 20, and a repetition penalty of 1.05. The maximum number of actions is set to 50.

Web search and page browsing are implemented using the Google Serper API and Jina Reader API, respectively. The VQA tool is based on Qwen2.5-VL-32B-Instruct~\cite{Qwen2.5-VL}, which takes a question and an image as input and outputs a model-generated response. Tool retrieval is performed using bge-large-en-v1.5~\cite{C-Pack}. All tool documentation follows the standard OpenAI function definition format: {\{"name": "...", "description": "...", "parameters": \{"type": "object", "properties": \{"param1": \{"type": "...", "description": "..."\}, ..., "required": ["param1"]\}\}}. This format is used for building the toolset index and for all prompts given to the agents.

Training consists of 100 steps of ToolPO with a batch size of 64, $\lambda_1=\lambda_2=1$, rollout size $K=8$, and a maximum sequence length of 32,768. The maximum number of actions is 50. The training framework is based on VeRL~\cite{verl} for multi-node distributed training. All experiments are conducted on 64 NVIDIA H20-141GB GPUs.

\section{Memory Schema}
\label{app:memory_schema}
Our brain-inspired memory architecture contains three components: episodic, working, and tool memory. To support stable memory folding and reduce information loss, we define each component with a fixed JSON schema, enabling reliable parsing and use of compressed memories for long-horizon reasoning.

\paragraph{Episodic Memory Schema}
Episodic memory records high-level task progression (milestones, decisions, outcomes) to preserve long-term context. The format is:
{\{"task\_description": "A general summary of what the reasoning history has been doing and the overall goals it has been striving for.", "key\_events": [\{"step": "step number", "description": "A detailed description of the specific action taken, decision made, or milestone achieved at this step, including relevant context and reasoning behind the choice.", "outcome": "A detailed account of the direct result, observation, or feedback received from this action or decision, including any new information gained or changes in the task state."\}], "current\_progress": "A general summary of the current progress of the task, including what has been completed and what is left to be done."\}}

\paragraph{Working Memory Schema}
Working memory captures the immediate goal, active challenges, and next actions to maintain continuity across folds. The format is:
{\{"immediate\_goal": "A clear summary of the current subgoal—what you are actively working toward at this moment.", "current\_challenges": "A concise summary of the main obstacles or difficulties you are presently encountering.", "next\_actions": [\{\"type": "tool\_call or planning ordecision", \"description": "Anticipate and describe the next concrete action you intend to take to advance the task."\}]\}}

\paragraph{Tool Memory Schema}
Tool memory consolidates tool-use experience (success rates, effective parameters, common errors) and derives reusable rules. The format is:
{\{"tools\_used": [\{"tool\_name": "string", "success\_rate": "float", "effective\_parameters": ["param1", "param2"], "common\_errors": ["error\_type1", "error\_type2"], "response\_pattern": "description of typical output", "experience": "Reflect and summarize your experience, including both successes and failures."\}], "derived\_rules": ["When X condition occurs, prefer tool Y", "Tool Z works best with parameter A set to B"]\}}

\section{Case Study}
\label{app:case_studies}
To illustrate the effectiveness of our DeepAgent framework in handling complex, multi-step tasks that require coordinated use of multiple tools, we present a detailed case in Table~\ref{tab:case_deepagent}. This example demonstrates how DeepAgent autonomously navigates tool selection, executes sequential actions, and synthesizes results to provide comprehensive solutions to user queries.

\begin{table}[t]
\setlength{\abovecaptionskip}{0.2cm}
\setlength{\belowcaptionskip}{0cm}
\centering
    \caption{A case study of DeepAgent for multi-step, multi-tool tasks on ToolBench, showing only key actions.}
    \fontsize{8pt}{9.6pt}\selectfont
    \begin{tabular}{p{0.95\columnwidth}}
    \toprule
        \rowcolor{gray!20}\textbf{Example of DeepAgent on a Multi-Tool Task} \\
    \midrule
        \textbf{User Query:} \\
        I'm organizing a film festival and I need assistance in finding the best films. Can you search for videos related to 'documentary' on Vimeo? Additionally, fetch the related people in the 'cinema' category to invite them as guest speakers. Finally, provide me with a streaming link for a YouTube video with the ID 'UxxajLWwzqY'. \\
        \midrule
        \rowcolor{gray!20}
        \textbf{DeepAgent Processing Output}\\
    \midrule
        Okay, I need to solve three tasks: find documentaries on Vimeo, find cinema-related people on Vimeo, and get a YouTube streaming link. I'll start by searching for the right tools.
        \\
        \green{<tool\_search>} Vimeo video search for documentaries \green{</tool\_search>}
        The search returned several tools. {search\_videos} and {get\_videos\_by\_tag} seem most relevant. I'll use the search tool first to find documentaries.
        \\
        \purple{<tool\_call>} \{"name": "search\_videos", "arguments": \{"query": "documentary", "sort": "most\_liked"\}\} \purple{</tool\_call>}
        The tool returned a list of popular documentary videos. Now for the second task: finding people in the 'cinema' category. Since there is no direct tool to search for people, I will try searching for videos with the 'cinema' tag and identify their creators.
        \\
        \purple{<tool\_call>} \{"name": "get\_videos\_by\_tag", "arguments": \{"tag": "cinema", "sort": "most\_liked"\}\} \purple{</tool\_call>}
        This call returned videos tagged with 'cinema' and included user information for each video, such as "John Doe" and "Jane Smith", who are potential guest speakers.
        \\
        Finally, for the YouTube streaming link, I'll first verify the video ID exists using the YouTube search tool.
        \\
        \purple{<tool\_call>} \{"name": "search\_for\_youtube\_v3", "arguments": \{"part": "snippet", "type": "video", "q": "UxxajLWwzqY"\}\} \purple{</tool\_call>}
        The tool confirmed the video ID is valid. I can now construct the streaming link. I will now consolidate all the information to answer the user's query.
        \\
        \textbf{Final Answer:} \\
        1. Documentaries on Vimeo: Found nature and space documentaries.\\
        2. Cinema Guest Speakers: John Doe, Jane Smith with Vimeo profiles.\\
        3. YouTube Link: https://www.youtube.com/watch?v=UxxajLWwzqY \\
    \bottomrule
    \end{tabular}
    \label{tab:case_deepagent}
\end{table}

\end{CJK}
\end{document}